\documentclass[journal]{IEEEtran}
\usepackage{ifpdf}
\usepackage{amsmath}
\usepackage{amsfonts}
\usepackage{algorithmic}
\usepackage{hyperref}
\usepackage{array}
\usepackage{url}
\usepackage{enumitem}
\usepackage{wrapfig}
\usepackage{cite}
\usepackage{comment}
\usepackage{xcolor}
\usepackage{float}

\newcommand{\upd}[1]{\textcolor{black}{#1}}
\newcommand{\upr}[1]{\textcolor{black}{#1}}

\ifCLASSOPTIONcompsoc
  \usepackage[caption=false,font=normalsize,labelfont=sf,textfont=sf]{subfig}
\else
  \usepackage[caption=false,font=footnotesize]{subfig}
\fi

\ifCLASSINFOpdf
  \usepackage[pdftex]{graphicx}
\else
  \usepackage[dvips]{graphicx}
\fi

\begin{document}
\bstctlcite{IEEEexample:BSTcontrol}
%
\title{Learning a Generative Motion Model from Image Sequences based on a Latent Motion Matrix}
%
%
%
\author{Julian Krebs, Herv\'{e} Delingette, Nicholas Ayache and Tommaso Mansi

\thanks{This work has been supported by the French government, through the 3IA C\^{o}te d'Azur Investments in the Future project managed by the National Research Agency (ANR) with the reference number ANR-19-P3IA-0002 and the grant AAP Sant\'e 06 2017-260 DGA-DSH, and by the Inria Sophia Antipolis - M{\'e}diterran{\'e}e, "NEF" computation cluster. The used data were obtained from the EU FP7-funded project MD-Paedigree and the ACDC STACOM challenge 2017 \cite{bernard2018deep}.}
\thanks{J. Krebs is with the Universit\'{e} C\^{o}te d'Azur, Inria, Epione Team, Sophia Antipolis, 06902 France and also Siemens Healthineers, Digital Technology \& Innovation, Princeton, NJ 08540 USA (e-mail: julian.krebs@inria.fr).}%
\thanks{H. Delingette and N. Ayache are with the Universit\'{e} C\^{o}te d'Azur, Inria, Epione Team, Sophia Antipolis, 06902 France.}
\thanks{T. Mansi is with Siemens Healthineers, Digital Technology \& Innovation, Princeton, NJ 08540 USA.}
}



\maketitle

\begin{abstract}
We propose to learn a probabilistic motion model from a sequence of images for spatio-temporal registration. Our model encodes motion in a low-dimensional probabilistic space -- the motion matrix -- which enables various motion analysis tasks such as simulation and interpolation of realistic motion patterns allowing for faster data acquisition and data augmentation. More precisely, the motion matrix allows to transport the recovered motion from one subject to another simulating for example a pathological motion in a healthy subject without the need for inter-subject registration. 
The method is based on a conditional latent variable model that is trained using amortized variational inference. This unsupervised generative model follows a novel multivariate \upd{Gaussian process} prior and is applied within a temporal convolutional network which leads to a diffeomorphic motion model. Temporal consistency and generalizability is further improved by applying a temporal dropout training scheme. Applied to cardiac cine-MRI sequences, we show improved registration accuracy and spatio-temporally smoother deformations compared to three state-of-the-art registration algorithms. Besides, we demonstrate the model's applicability for motion analysis, simulation and super-resolution by an improved motion reconstruction from sequences with missing frames compared to linear and cubic interpolation.
\end{abstract}


%
\IEEEpeerreviewmaketitle

\section{Introduction}
\IEEEPARstart{M}{otion} analysis is an important task in many medical image analysis problems such as organ tracking or longitudinal analysis of various diseases. For moving organs such as the heart, it is not only important to track anatomical structures but also to analyze motion indices that are useful for disease diagnosis or therapy selection \cite{girija20174d}. Extracting motion patterns further allows to compensate for motion, handle missing data or do temporal super-resolution and motion simulation.

Motion in medical image sequences is typically analyzed by computing temporally consistent pairwise deformations where each frame in a sequence is registered to a target frame \cite{girija20174d}. The resulting series of deformation fields can be utilized to track structures throughout the sequence and to identify abnormal motion patterns, for example by computing clinically relevant variables such as the ejection fraction (EF) of the heart \cite{rohe2018low}. 

\subsection{State-of-the-art}
Registration algorithms typically seek to find the deformation field between two images by solving an optimization problem consisting of a similarity metric and a regularizer. The similarity metric measures the distance between the two images while the regularizer constrains the smoothness of the resulting deformation field. A large variety of registration algorithms using different similarity and regularizing metrics have been proposed  \cite{sotiras}. One group of registration methods \upr{aims} to ensure diffeomorphic deformations due to their favorable properties. Diffeomorphisms are topology-preserving and invertible deformations which makes them suitable for many medical registration problems in which foldings are physically implausible \cite{vercauteren2009diffeomorphic}. This makes diffeomporphisms also appropriate for tracking anatomical structures in image sequences such as in cardiac imaging \cite{peyrat2010registration} (assuming structures do not go out of the field of view). Many diffeomorphic registration algorithms have been proposed such as  \cite{beg2005computing,zhang2015finite,vercauteren2008symmetric, vercauteren2009diffeomorphic}, the SyN algorithm \cite{avants2008symmetric} and the LCC-demons \cite{lorenzi2013lcc}. 
Recently, learning-based algorithms for pairwise diffeomorphic registration have been proposed. These are based on supervised \textit{ground-truth} deformations \cite{yang2017quicksilver,rohe2017svf} or on unsupervised learning \cite{dalca2018unsupervised,krebs2019learning}. The latter are trained by minimizing a loss function consisting of an image similarity and a deformation regularizer, similarly to the traditional optimization problem, which has been proposed earlier for learning-based non-diffeomorphic registration \cite{de2019deep,dalca2019unsupervised}. In \cite{dalca2018unsupervised,krebs2019learning}, diffeomorphisms are guaranteed by using the stationary velocity field (SVF) parameterization based on the scaling-squaring algorithm \cite{arsigny2006log}.

For image sequences, one difficulty is to acquire temporally smooth deformations that are fundamental for consistent tracking. That is why registration algorithms with a temporal regularizer have been proposed \cite{ledesma2005spatio,vandemeulebroucke2011spatiotemporal,de2012temporal,metz2011nonrigid,qin2018joint,shi2013temporal}. \upd{For respiratory motion modeling, learning-based regression models using ultrasound and MR images of different respiratory states have been used to learn respiratory motion patterns \cite{giger2018respiratory,giger2019inter}.} 
In the computer vision community, temporal video super-resolution and motion compensation are a related research topic \cite{caballero2017real,kappeler2016video}.

However, while these methods are able to capture temporally consistent deformations along a sequence of images, they do not extract intrinsic \upd{low-dimensional} motion parameters crucial for building a comprehensive motion model that can be used for analysis tasks such as motion simulation, transport or classification as it is for example done in bio-mechanical models such as \cite{sermesant2008toward}. Yang et al.~\cite{yang2011prediction} generated a motion prior using manifold learning from low-dimensional shapes. Qiu et al.~\cite{qiu2011principal} proposed to build an eigenspace of initial momenta using PCA. In an image-driven fashion, Roh\'{e} et al.~\cite{rohe2018low} introduced  a  parameterization, the Barycentric Subspaces, for cardiac motion analysis. \upd{It has been also proposed to learn statistical models for respiratoy motion by combining sample transformations derived from shapes \cite{jud2015respiratory} or images directly \cite{jud2017localized}. However, these models require either to extract shapes from images and/or an inter-subject alignment which can be a difficult task with regard to different use cases.}

\subsection{Learning a Probabilistic Motion Model}
In contrast, we propose a probabilistic motion model that is built in a fully data-driven way from image sequences. \upd{Our model learns a low-dimensional motion matrix in an unsupervised fashion from sequences with constant intensity levels. Instead of defining a motion parameterization explicitly or learning from pre-processed shapes or pre-aligned sequences, an application-specific motion model is learned. The advantage of such an application-specific model is that it is guided from training images which allows to unveil inherent characteristic motion features instead of for example relying on pre-defined bio-mechanical parameters often based on limiting assumptions.} The goal is not only to retrieve a compact representation of the motion but to obtain a structured and generative encoding that allows for temporal interpolation (to predict missing frames) and to simulate an indefinite number of new motion patterns. These features could be helpful for data augmentation and to speed-up image acquisition as the model reconstructs a full cyclic motion from missing frames. \upd{As for all learning-based approaches, our resulting model is biased on the training data which mostly impacts its generative abilities. Naturally, it will tend to simulate pathological cases if trained mostly on image sequences containing similar pathologies.} Besides, \upd{the application-specific} probabilistic encoding could be useful for group-wise analysis as it enables to transport motion characteristics to a new subject \upd{-- without requiring inter-subject alignment --} simulating for example a pathological motion in a healthy subject. \upd{This can be useful for data augmentation and class balancing for instance by generating many simulated examples of a certain disease.}

In this work, we introduce a novel Gaussian process (GP) prior to extend a conditional variational autoencoder (CVAE \cite{kingma2014semi}), a latent variable model, for temporal sequences. \upr{The GP prior constrains the standard independence assumption between all variables of CVAEs with a temporal regularizer. Relating latent variables over time leads to higher temporal consistency that can improve the tracking of for example moving organs.} A pairwise encoder-decoder neural network applies a temporal convolutional network (TCN) in its latent space in order to learn intrinsic temporal dependencies. Furthermore, we utilize a self-supervised training scheme based on temporal dropout (TD) to enforce temporal consistency and increase generalizability of the motion model. Smooth and diffeomorphic deformations are guaranteed by applying an exponentiation layer \cite{krebs2019learning} and spatio-temporal regularization. 

The proposed model demonstrates state-of-the-art registration accuracy measured on segmentation overlaps and distances and regularity for diffeomorphic tracking of cardiac cine-MRI. In addition, the potentials of the generated latent motion matrix for motion simulation, interpolation and transport are demonstrated. The main contributions are as follows:
\begin{itemize}[noitemsep]
\item An unsupervised probabilistic motion model learned from medical image sequences
\item A conditional VAE model trained with a novel Gaussian process prior and self-supervised temporal dropout using temporal convolutional networks
\item Demonstration of cardiac motion tracking, simulation, transport and temporal super-resolution
\end{itemize}

This paper extends our preliminary conference paper \cite{krebs2019probabilistic} by replacing the standard unit Gaussian of the CVAE with a novel \upr{Gaussian process prior}. We add detailed derivations of the motion model and show improved tracking accuracy and temporal smoothness. Finally, we show a first generalization of the model to \upd{3D+t} sequences.

\section{Methods}
\upd{In this work, motion is described by deformation fields between one reference image, for example the first frame, and all other images in an image sequence $I_{0:T}$ with $T+1$ frames.} In order to extract consistent sequential deformations $\phi_t$ with $t\in [1,T]$, we propose a temporal latent variable model that encodes the motion in a low-dimensional probabilistic space, the motion matrix $z\in\mathbb{R}^{D\times \upd{T}}$.
Here, we define the reference image $I_0$ as moving image, while the other frames are fixed images $I_t$. Each image pair $(I_0, I_t)$ is encoded by $D$ latent variables, the \upr{$z_{\cdot t}$}-code, which are the columns of $z$. Each \upr{$z_{\cdot t}$} parameterizes the deformation field $\phi_t$ while being conditioned on the moving image $I_0$. The rows \upr{$z_{d\cdot}$} with length $\upd{T}$ of the motion matrix $z$ represent the encoded deformation sequence per latent dimension $d\in D$.

Our motion model is learned from data by imposing a Normal prior distribution $p(z)$ on the latent variables $z$ that follows a Gaussian process (GP) prior in the temporal dimension for each \upr{$z_{d\cdot}$}. In addition, we assume independence between the latent variables \upr{$z_{d\cdot}$} as in standard VAEs \cite{kingma2013auto}. Note, when $z$ is written as part of a distribution like $p(z)$, $z$ is used as a vector of size $D\upd{T}$ \upr{in row-major order} rather than a matrix for simpler notation. 

During training, we follow the learning paradigms of conditional variational autoencoders (CVAE \cite{kingma2014semi,kingma2014adam}) with the exception of replacing the multivariate unit Gaussian prior with the proposed GP-prior. The approximated posterior is the output of a temporal convolutional neural network (TCN \cite{koltun2018}) allowing for temporal regularization. To further facilitate temporal dependencies and handle missing data, temporal dropout (TD) is applied during the training procedure. In the following, the different parts of the method are explained. First, the probabilistic motion model using a GP-prior is defined. Then, posterior and data likelihood distributions are modeled using a encoder-decoder neural network. Lastly, the concept of temporal dropout is introduced. 

\begin{figure}[tb]
\centering 
\subfloat[]{\includegraphics[trim=5 295 650 20,clip,width=0.55\linewidth]{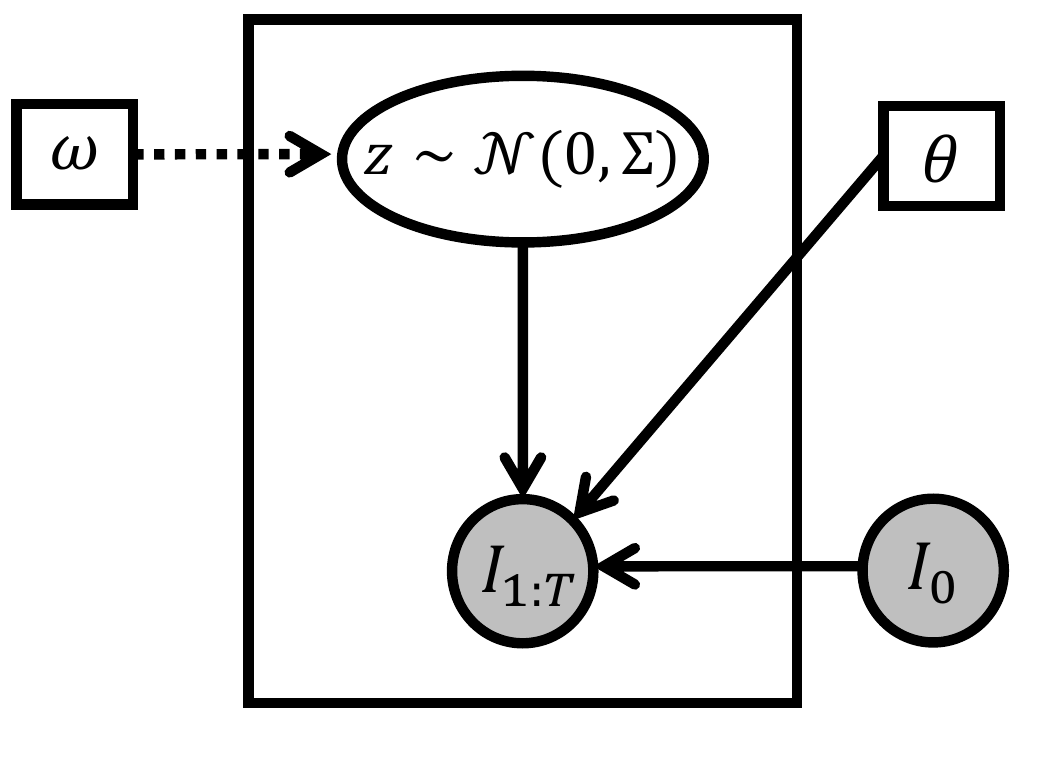}\label{motion:stoch_proc_figure}}\hfill
\subfloat[]{\includegraphics[trim=45 185 560 8,clip,width=0.45\linewidth]{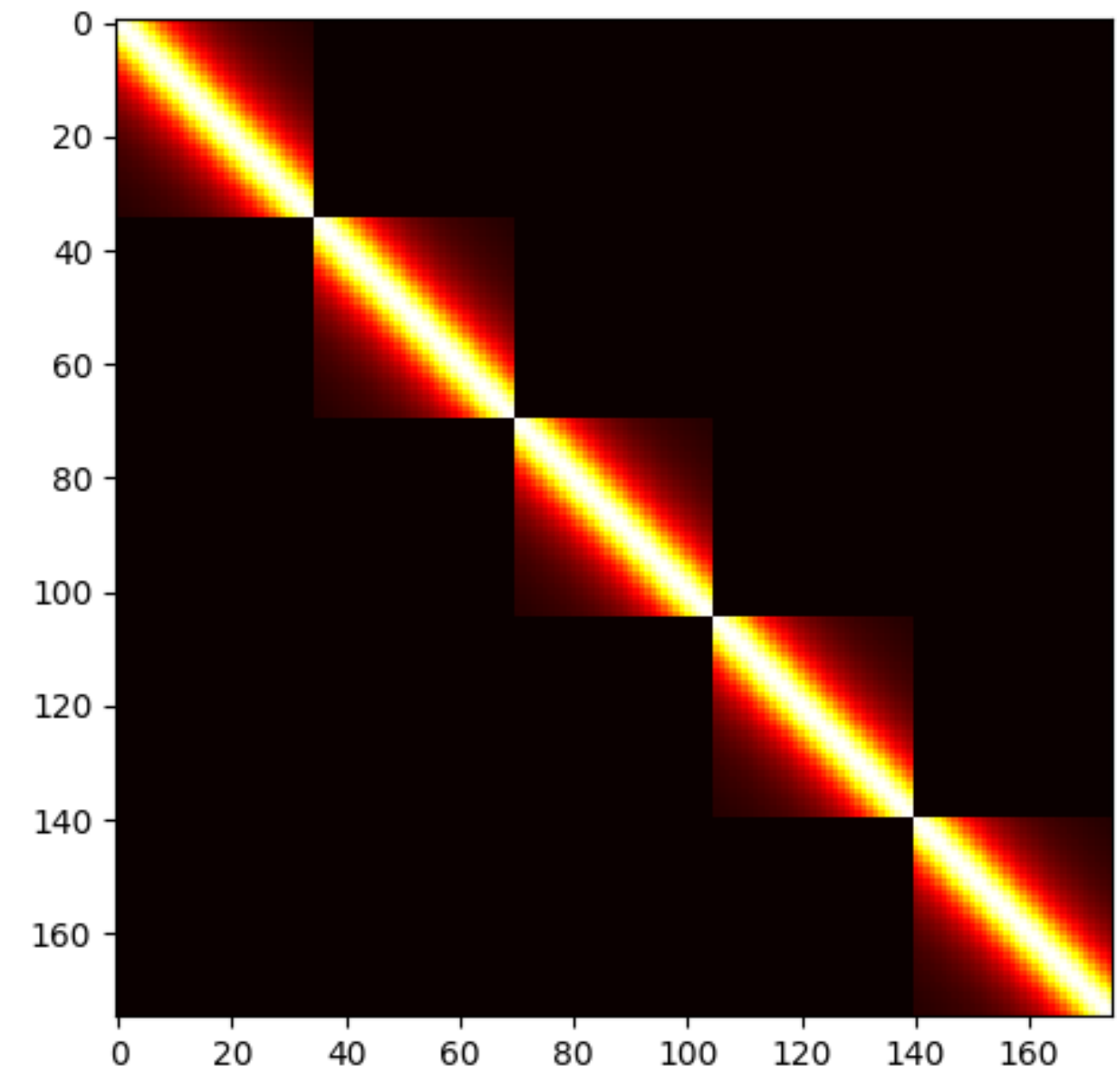}\label{motion:cov_prior_example}}
\caption{\small{(a) Generative process for the motion model representing the likelihood of fixed images $I_{1:T}$ given the latent variables $z$ and moving image $I_0$: $p_\theta(I_{1:T}|z,I_0)$, where $\omega$ and $\theta$ are fixed parameters and arrows denote dependencies between random variables. (b) Visualization of the covariance matrix $\Sigma$ of the Gaussian prior $p(z)$ with 5 latent dimensions, a sequence time length of 35 and a length scale of the Cauchy kernel of 7.}}
\end{figure}

\begin{figure}[tb]
\centering 
\includegraphics[trim=0 186 690 0,clip,width=1\linewidth]{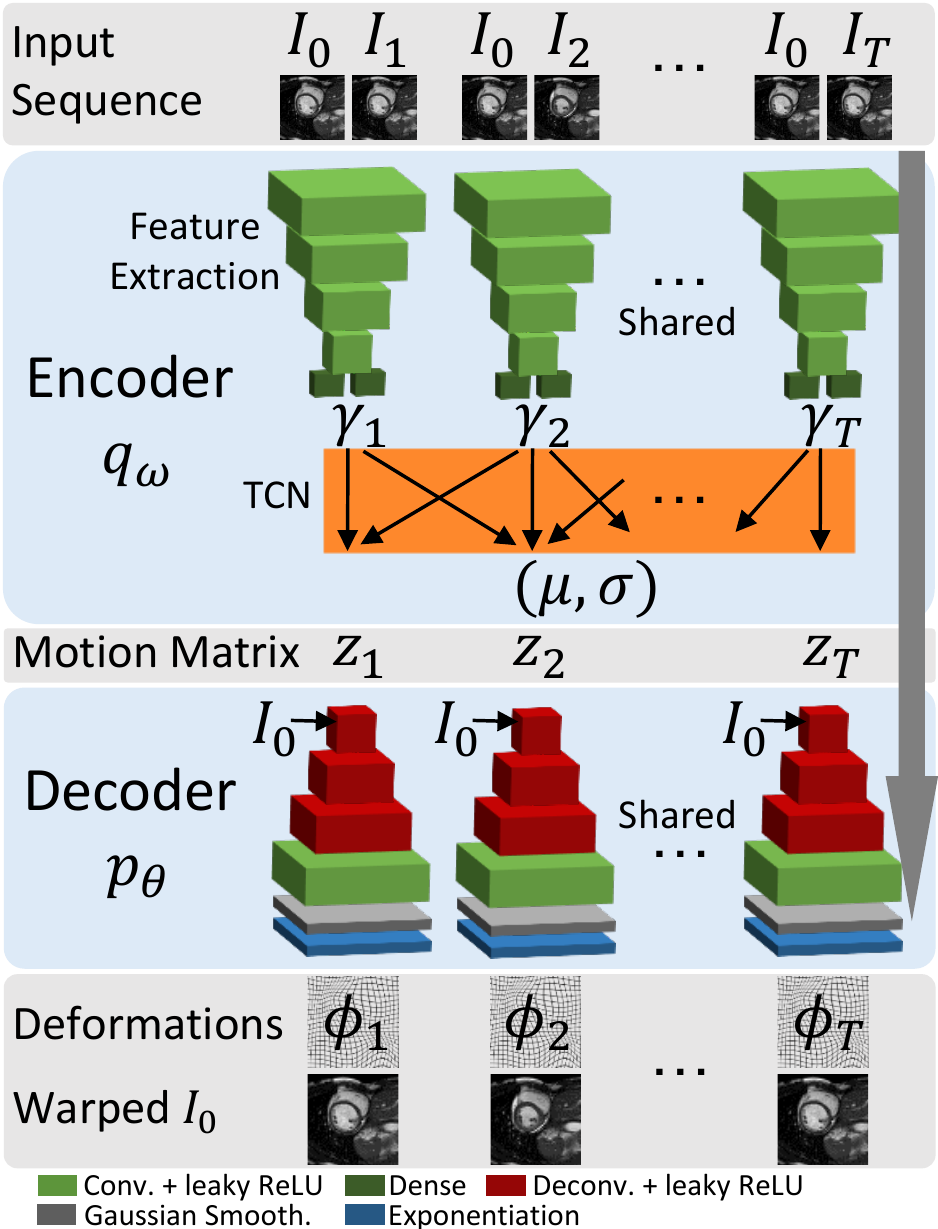}
\caption{\small{Overview of the motion model including encoder and decoder neural networks. From sequential image pairs, temporally independent feature vectors $\gamma_t$ are extracted which are fed to a temporally convolutional network (TCN) to obtain the probabilistic motion matrix $z$. This compact representation is decoded to a sequence of diffeomorphic deformation fields $\phi_t$.}}\label{motion:overview}
\end{figure}

\subsection{Generative Motion Model using a Gaussian Process Prior} 
The proposed motion model consists of an encoder $q_\omega(z|I_{0:T})$ and a decoder $p_\theta(I_{1:T}|z,I_0)$ which are parameterized by $\omega$ and $\theta$ respectively. The encoder first independently maps each image pair $(I_0, I_t)$ to a pair-wise latent representation which is then temporally regularized by mixing all time steps \upd{using multiple temporal 1D convolutions (TCN) to retrieve a joint latent representation $(\mu, \sigma)$. The motion matrix $z$ is finally extracted by sampling from the posterior distribution $q_\omega$ which is defined as a Normal distribution that is parameterized by $(\mu, \sigma)$. On the other hand, the} decoder $p_\theta$ projects the \upr{$z_{\cdot t}$}-codes to the \upr{deformation field} $\phi_t$ while being conditioned on the moving image $I_0$. The output of the decoder are \upd{the} reference image $I_0$ warped with the $\phi_t$ deformation fields. The decoder represents data likelihood of the latent variable model. Using a prior distribution $p(z)$ over latent variables $z$, we define the following generative process:
\begin{equation}
p_\theta(I_{1:T}|I_0)=\int_z p_\theta(I_{1:T}|z,I_0) p(z)~dz,
\end{equation}
which is visualized in Fig.~\ref{motion:stoch_proc_figure}.
In this work, encoder $q_\omega$ and decoder $p_\theta$ are approximated using neural networks where $\omega$ and $\theta$ represent the encoder and decoder networks' weights which are optimized using amortized Variational Inference \cite{kingma2013auto}. The data likelihood $p_\theta(I_{1:T}|z,I_0)$ can be seen as the fidelity of the reconstruction of the fixed images $I_{1:T}$ by warping the moving image $I_0$ with appropriate deformations $\phi_{1:T}$. An overview of the motion model can be seen in Fig.~\ref{motion:overview}.
\subsubsection{\upr{Gaussian process prior}} The prior follows a zero-centered multivariate Gaussian distribution: \upr{$p(z)\sim \mathcal{N}(0,\Sigma)$} where the covariance matrix $\Sigma$ is a diagonal block matrix of dimensions $D\upd{T}\times D\upd{T}$:
\begin{equation}\label{motion:cov_prior}
\Sigma=\text{Diag}_{d=1}^D (K_l).
\end{equation}
Each diagonal element of $\Sigma$ represents the temporal covariance matrix $K_l\in \mathbb{R}^{\upd{T} \times \upd{T}}$ of a Gaussian time-continuous stochastic process whose kernels can be chosen by the user. A typical choice in Gaussian processes is the squared exponential kernel $K_l^\text{RBF}(\tau,\tau')=\sigma_K^2\exp{(-|\tau-\tau'|^2/2l^2 ) }$ with length-scale $l$ and variance $\sigma_K^2$. \upr{The length-scale practically describes, how close two points $\tau$ and $\tau’$ have to be to influence each other significantly. Therefore, it represents a number or range of time steps.} \upd{Due to the fact that we want to model data that changes slowly and consistently over time we consider the Cauchy kernel \cite{rasmussen2003gaussian,fortuin2019multivariate} that is heavy-tailed and has a long-range influence:}
\begin{equation}\label{motion:kernel}
K_l^\text{Cauchy}(\tau,\tau')=\sigma_K^2\bigg( 1-\frac{(\tau-\tau')^2}{l^2}\bigg)^{-1},
\end{equation}
with pre-defined $\sigma_K$. This covariance matrix $\Sigma$ allows temporally correlated latent variables while still assuming highest possible independence between the $D$ latent dimensions. In other words, we extended the standard VAE latent space which only consists of the independence assumption between latent variables with a regularized temporal dimension. Latent variables are related over time according to the chosen kernel function $K_l$ while being independent of each other. An example of a  covariance matrix can be seen in Fig.~\ref{motion:cov_prior_example}.
\subsubsection{Posterior and Likelihood Distributions} Similar to standard VAEs, the posterior $q_\omega$ follows a multivariate Gaussian distribution \upr{$q_\omega(z|I_{0:T})\sim \mathcal{N}(\mu , \Sigma^*(\sigma) )$} with data-driven predictions of mean vector $\mu \in \mathbb{R}^{D\upd{T}}$ and variance vector $\sigma \in \mathbb{R}^{D}$. The full covariance matrix $\Sigma^*(\sigma)$ is defined as a \upd{$D$-dimensional} block diagonal matrix of the following form \upd{where each block is scaled using the values of the variance vector $\sigma$:
\begin{equation}\label{motion:cov}
\Sigma^*(\sigma) = \begin{bmatrix}\sigma_1 K_l &0 &\cdots &0 \\0 &\sigma_2 K_l &\cdots &0 \\ 
\vdots & \vdots & \ddots & \vdots\\
0 &0 &\cdots &\sigma_D K_l
\end{bmatrix}.
\end{equation}}
Mean and variance vectors $(\mu, \sigma)$ are the output of the encoder neural network. The kernel $K_l$ is kept the same as in the prior distribution and does not contain predicted parameters to guarantee a user-chosen temporal regularity.

Also, the likelihood $p_\theta$ is assumed to follow a multivariate Gaussian distribution \upr{$p_\theta(I_{1:T}|z,I_0) \sim \mathcal{N}(I_0 \circ  \phi_{1:T}(\theta) , \sigma_L \ast I_{D\upd{T}})$} where \upd{the mean is represented by the warped moving image $I_0$ that is deformed by the diffeomorphisms $\phi_{1:T}$ (the decoder output) and $\circ$ denotes the image warping operation. The covariance is represented by the identity matrix $I_{D\upd{T}}$ of size $D\upd{T}$ times (denoted by $\ast$) the scalar $\sigma_L$ which can depict for example the variance of intensity residuals of well registered images.}
\subsubsection{Learning the Motion Model via Variational Inference}
In order to optimize the parameterized motion model over $\omega$ and $\theta$, the evidence lower bound (ELBO) of the log-marginalized likelihood $p_\theta(I_{1:T}|I_0)$ that is conditioned on the moving image $I_0$, must be maximized (see \cite{kingma2013auto,kingma2014semi,krebs2019learning} for details):
\upd{\begin{multline}\label{motion:elbo}
\log{p_\theta(I_{1:T}|I_0)} - \text{KL}\big[ q_\omega(z|I_{0:T}) \| p(z|I_{0:T})\big] = \\
\mathbb{E}_{z\in q_\omega (\cdot | I_{0:T})}\big[\log{p_\theta(I_{1:T}|z,I_0)} \big] - \text{KL}\big[ q_\omega(z|I_{0:T}) \| p(z)\big],
\end{multline}}
with KL denoting the Kullback-Leibler Divergence (KL). The first term in \ref{motion:elbo} enforces that the moving image $I_0$ is well registered to the fixed images $I_{1:T}$ by maximizing the log likelihood. The second term structures the latent motion encoding by enforcing the posterior distribution $q_\omega(z|I_{0:T})$) to be close to the prior distribution $p(z)$. Following the definition of the KL divergence between 2 multivariate Gaussian distributions, we obtain the closed-from solution (see Appendix A):
\begin{equation}\label{motion:kl}
\text{KL}\big[ q_\omega(z|I_{0:T}) \| p(z)\big]= \frac{1}{2}\sum_{i=1}^D \sigma_i^2\upd{T} + \bar{\mu}_i^\top K^{-1} \bar{\mu}_i - \log{(\sigma_i^2)}-\upd{T},
\end{equation}
with $\bar{\mu}_i$ being the $i$-th segment of length $T$ in $\mu$.

Recall that the log likelihood $p_\theta(I_{1:T}|z,I_0)$ is also Gaussian. Thus, \upd{$\log{p_\theta(I_{1:T}|z,I_0)} = -\frac{1}{2} \sum_{t=1}^T\| I_t - I_0 \circ \phi_t \|^2 / \sigma_L + C$ with a constant $C$} which is equivalent to adopting a sum-of-squared differences (SSD) criterion, commonly used as similarity metric in image registration (for example in \cite{balakrishnan2018unsupervised}).

During training of the model, parameters $\omega$ and $\theta$ are updated via stochastic gradient descent and back-propagation. In order to back-propagate through the sampling operation, the reparameterization trick is used \cite{kingma2013auto}. For full-covariance Gaussian distributions, the covariance matrix must be positive-definite as we use the Cholesky decomposition for the reparameterization (cf. \cite{kingma2019introduction}). The details on how to efficiently compute the Cholesky decomposition of the covariance matrix $\Sigma^*$ in \ref{motion:cov} can be found in Appendix B. \upd{Note that the covariance matrix grows quadratically with the sequence length. This could potentially lead to high computational expenses for longer image sequences. However, due to the diagonal block structure and the possibility of precomputing the Cholesky decomposition (Appendix B), computations are very efficient. Especially for common image sequence length in medical imaging. For sequences of 30-50 images, the covariance computations are computationally not relevant in comparison to the rest of the neural networks.}

\begin{figure}[tb]
\centering 
\subfloat[]{\includegraphics[trim=5 312 640 6,clip,width=0.47\linewidth]{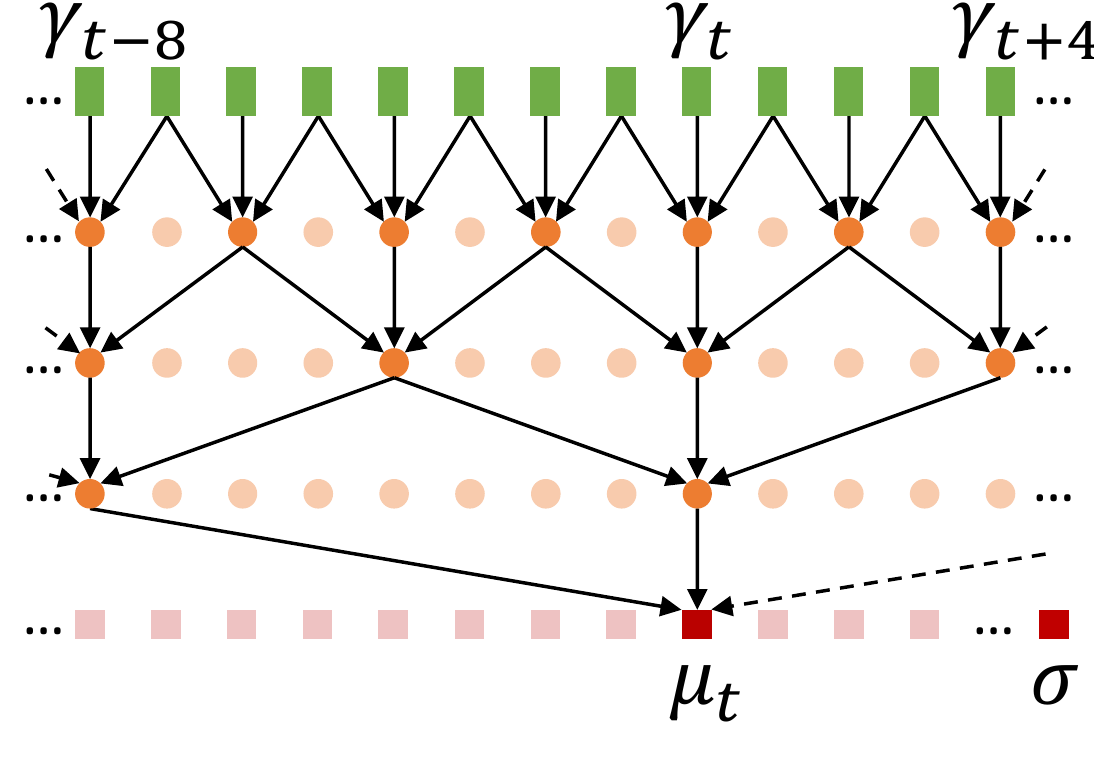}\label{motion:tcn}}\hfill
\subfloat[]{\includegraphics[trim=0 345 697 5,clip,width=0.5\linewidth]{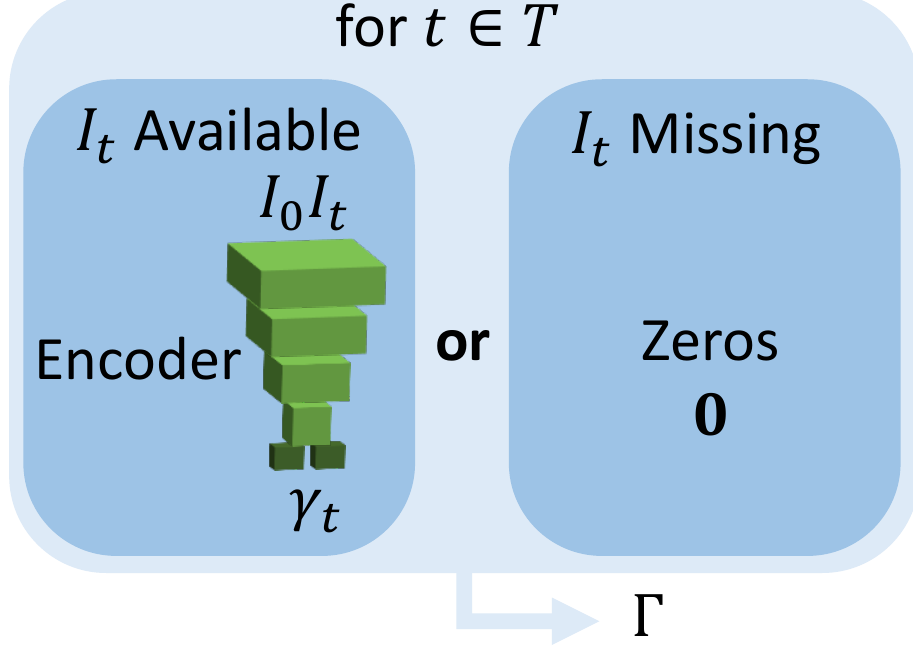}\label{motion:td}}
\caption{\small{(a) The temporal convolutional network (TCN) allows for temporal regularization of the independently extracted features $\gamma_t$ per time step $t$, for retrieving mean vector $\mu$ and variance vector $\sigma$ of the posterior distribution $p_\theta$. (b) Sequences with missing time steps (motion interpolation or simulation) are encoded by a full feature matrix $\Gamma$ by setting the columns of missing time steps to zero. The TCN handles these missing columns and still predicts a full temporal motion sequence of $\upd{T}$ time steps.}}
\end{figure}

\subsubsection{Neural Network Architecture}
\upr{A graphical summary of the temporal encoder-decoder neural network architecture is shown in Fig.~\ref{motion:overview}.} The encoder takes the image pairs $(I_0,I_t)$ as input and outputs the motion matrix $z$. It consists of a feature extraction part and a temporal regularizer (TCN). The feature extraction part consists of \upd{\upr{three down-sampling and one size-maintaining convolutional layers} and one fully-connected layer of size $2D$ for \upr{low-dimensional} mean and variance predictions of the posterior \cite{kingma2014semi}. As non-linearities, leaky rectifier functions \cite{maas2013rectifier} have been used in the convolutional layers while the fully-connected layer is linear. These layers are temporally independent and share weights across all image pairs of a sequence. The output of the feature extraction networks is the feature matrix $\Gamma$ with feature vectors $\gamma_t$ per time step $t$ of size $\mathbb{R}^{2D}$. These feature vectors are temporally regularized by merging them across different time steps using a temporal convolutional network (TCN). This leads to temporally consistent mean and variance vectors $(\mu,\sigma)$ that define the posterior distribution \upr{$q_\omega(z|I_{0:T}) \sim \mathcal{N}(\mu , \Sigma^*(\sigma) )$}.} The size of $2D$ is chosen for $\gamma_t$ such that each $\sigma$ value can be influenced by features from the whole sequence. Note, that samples from the posterior distribution are vectors of size $D\upd{T}$ which are reshaped to retrieve the motion matrix $z$ with \upr{$z_{\cdot t}$}-columns. 

Following the recommended architecture, the TCN consists of 1D convolutional layers with increasing dilation and skip connections allowing to learn temporal dependencies of the latent variables $\gamma_t$ that were time-independent before \cite{koltun2018}. We use zero-padding and non-causal convolutional layers to also take future time steps into account. \upd{In particular, four consecutive blocks of convolutions are used each consisting of a 1D convolution with 3x3 filters with rectifier non-linearities, a spatial 1D dropout layer and followed by a 1x1 linear convolution layer. In addition each block's output is added to their input to establish skip connections. An additional 1x1 linear convolutional layer is used as the first TCN layer and the output of the TCN is the sum of the outputs of all blocks. Size-maintaining zero-padding is used and the number of filters is $2D$ for all TCN layers to keep input and output matrices of same dimensions.  The output tensor of size $\mathbb{R}^{D\upd{T}+D}$ is split into $\mu$ and $\sigma$ vectors where exponentiation is applied on the $\sigma$-vector to guarantee non-negative values close to 1.} Our TCN is shown in Fig.~\ref{motion:tcn}. TCNs can handle  sequences of varying time lengths and are advantageous compared to recurrent neural networks (RNN) due to a flexible receptive field and more stable gradient computations \cite{koltun2018}. Another reason why the authors chose a TCN over RNNs is that RNNs are especially suitable to learn long-distance temporal relationships such as in natural language processing while the focus of this work is on rather short time sequences with higher local dependencies. One could use a cyclic padding instead of zero-padding for cyclic sequences, for example by linking the end of a sequence to its beginning. However, in the case of cardiac cine-MRI, 5-10\% of the cardiac cycle are often omitted \cite{bernard2018deep} such that we chose to not assume cyclic sequences explicitly.

\upd{For each time step, the decoder takes as input the \upr{$z_{\cdot t}$} vector which are sampled from the posterior distribution by using the reparameterizaiton trick} and outputs the diffeomorphisms $\phi_{1:T}$ and the accordingly warped moving image. \upr{A fully-connected layer, three up-sampling} deconvolutional and \upd{two size-maintaining} convolutional layers are used in the decoder which are shared across all time steps. \upr{The $z_{\cdot t}$ is first extended and reshaped in order to fit the input size of the first deconvolutional layer.} It is desired that the latent representation $z$ encodes deformation information on a semantic level, independent of the given subject. That is why the decoder is further conditioned on the moving image $I_0$ by concatenating down-sampled versions of $I_0$ with the outputs of the deconvolutional layers at different scales. \upr{The image $I_0$ is hereby down-sampled by tri-linear interpolation with factors 2, 4 and 8 while the original sized $I_0$ is concatenated after the third deconvolutional layer.} By providing subject-specific appearance information in form of the moving image, the motion model is driven to encode subject-independent deformation information in the limited dimensionality of $z$ \cite{krebs2019learning}. \upd{Leaky rectifier functions \cite{maas2013rectifier} are used in the deconvolutional layer while a \textit{tanh} activation is applied after the last convolutional layer for stability reasons during training.} 

\upd{In addition, a diffusion-like regularization in spatial and temporal dimensions is applied by Gaussian smoothing kernels. This regularization follows the derivations of \cite{krebs2019learning} and is omitted in Fig.~\ref{motion:stoch_proc_figure} for reasons of clarity. We utilize a Gaussian smoothing layer with standard deviations of $\sigma_G$ and $\sigma_T$ in temporal and spatial domains respectively. To ensure diffeomorphic deformations, an exponentiation layer \cite{krebs2019learning} that relies on the \textit{scaling-squaring} algorithm \cite{arsigny2006log} for the stationary velocity field parameterization of diffeomorphisms  is applied. The differentiable} linear warping functionality is realized using a spatial transformer network layer \cite{jaderberg2015spatial}. \upd{The full details of architecture and training can be found in the implementation details section \ref{imp_deetails}.} \upr{Additionally, a table summarizing all the layers is presented in Appendix C.}

\subsection{Missing Data and Temporal Dropout}\label{motion:missingdata}
\upr{The temporal latent dimensionality $T$ (the size of the covariance matrix $\Sigma^*$) is kept identical across datasets with different time lengths $T^*$. However, the model can handle arbitrary lengths of sequences up to this maximum length $T$ which needs to be set before training starts (e.g. the maximum expected sequence length). } In case of shorter sequences, the features $\gamma_\tau$ of all available image pairs $(I_0, I_\tau)$ with $\tau \in T^*$ are extracted and evenly distributed along $T$ forming the matrix $\Gamma \in \mathbb{R}^{2D\times \upd{T}}$. The remaining missing time steps are filled with a constant (typically zero). \upd{As typical in handling of incomplete data in neural networks, in the course of training, this constant will be associated to as missing time steps as these values are not beneficial for optimizing the loss function.} On the decoder side, the log-likelihood loss (first part of \ref{motion:elbo}) is evaluated on all available time steps of the original sequence. If a sequence is longer than $T$, evenly distributed frames would be dropped to reach a  length of $T$. However, this should not happen normally as we assume to put $T$ at least as the maximum experienced length in the data.

In addition, during training, further time steps (i.e.~$\gamma_\tau$) are dropped from $\Gamma$ using temporal dropout (TD) in order to force the motion model to interpolate motion between available frames. To encourage the TCN to make use of its temporal connections and search for dependencies across time, our TC drops some of the $\gamma_\tau$ while still trying to recover the deformations $\phi_\tau$ of all available image pairs $(I_0,I_\tau)$. More precisely, in TD, instead of extracting features from an image pair $(I_0,I_\tau)$, a vector of zeros is chosen as $\gamma_\tau$ while still keeping the loss function on the decoder part for these time steps. A binary Bernoulli random variable $r_\tau$ is used to randomly choose at each original time step $\tau$ if the zero vector is used instead of the extracted features given $(I_0,I_\tau)$. All independent Bernoulli random variables $r\in \mathbb{R}^{T^*}$ have the success probability $\delta$. The latent feature representation $\gamma_t^{TD}$ using TD can thus be defined as:
\begin{equation}
\gamma_\tau^{TD} = r_\tau \ast \mathbf{0} + (1 - r_\tau) \ast \gamma_\tau.
\end{equation}
Note, TD is used only during training as a sort of self-supervision to encourage generalizability and consistent motion simulation and interpolation of missing data. When encountering missing data at test time, one just needs to place the available encoded frame pairs at the desired temporal positions of $\Gamma$ in order to predict the full motion consisting of $T$ time steps (cf. Fig.~\ref{motion:td}). A full motion simulation can be generated by setting all elements of $\Gamma$ to zero. In this case, a sequence of deformations that are plausible with respect to the training data will be predicted given only the original image $I_0$.

\textit{Optional Random Sub-Sequence Training:}
Since our motion model takes sequences of images as input and outputs a sequence of deformation fields, it comes naturally with high computational costs. This can lead to a model that may not be trainable on standard GPUs. Due to this limitation, we propose to train our model optionally with random sub-sequences. \upd{This can be done by dropping the encoder and decoder for some time steps while keeping the full temporal dimensionality in the latent space (the motion matrix and TCN).} Let $\mathcal{T}$ be the maximum number of frames with which our model can be trained on a given GPU. In each training iteration, a random combination of $\mathcal{T}$ frames is selected from a training subject with $T^*$ frames whenever $T^*>\mathcal{T}$. \upd{As in the case of missing data, the covariance matrix and TCN is kept with the original size containing $T$ time steps. The selected frame pairs are encoded and placed at their relative temporal position in $\Gamma$ while filling the remaining time steps with zeros. In contrast to the TD or motion interpolation procedure, only the selected $\mathcal{T}$ time steps are reconstructed in the decoder to limit the requirements of GPU memory.} In case of shorter training sequences with $T^*\leq\mathcal{T}$, the full sequence is used. By sampling different sub-sequences in each training epoch, the network will eventually see all parts of a sequence during the training stage.

\section{Experiments}
In this paper, we evaluate the proposed motion model on cardiac cine-MRI. Besides accurate temporal tracking and registration, we show the model's capabilities for motion simulation, interpolation and transport. The improved temporal latent space using the GP prior is demonstrated. Extensive results are presented for 2D+t sequences with more limited quantitative evaluations on 3D+t sequences due to their heavy computational requirements. In all experiments, the end-diastolic (ED) frame was used as the moving image $I_0$. 

\subsection{\upd{Data sets}}
Two data sets forming 334 cardiac cine-MRI in total were used. First, 184 multi-centric short-axis sequences came from the EU FP7-funded project MD-Paedigree (Grant Agreement 600932), with congenital heart disease and healthy or pathological images from adults. In addition, 150 sequences originated from the Automatic Cardiac Diagnosis Challenge 2017 (ACDC \cite{bernard2018deep}). The images were acquired in breath hold using 1R-R or 2R-R intervals mixing retrospective or prospective gating. The original sequence lengths varied from 13 to 35 frames. The 100 \textit{training} cases from ACDC that contain ED-ES segmentation information were used for testing while all other sequences were used for training. Slices were resampled with a spacing of 1.5$\times$1.5 mm and cropped to a size of 128$\times$128 pixels. In case of 3D+t sequences, 18 slices were used by adding zero slices at the top and bottom in case of fewer original slices.

\begin{figure*}[tb]
\centering
\includegraphics[trim=6 130 375 6,clip,width=1.\linewidth]{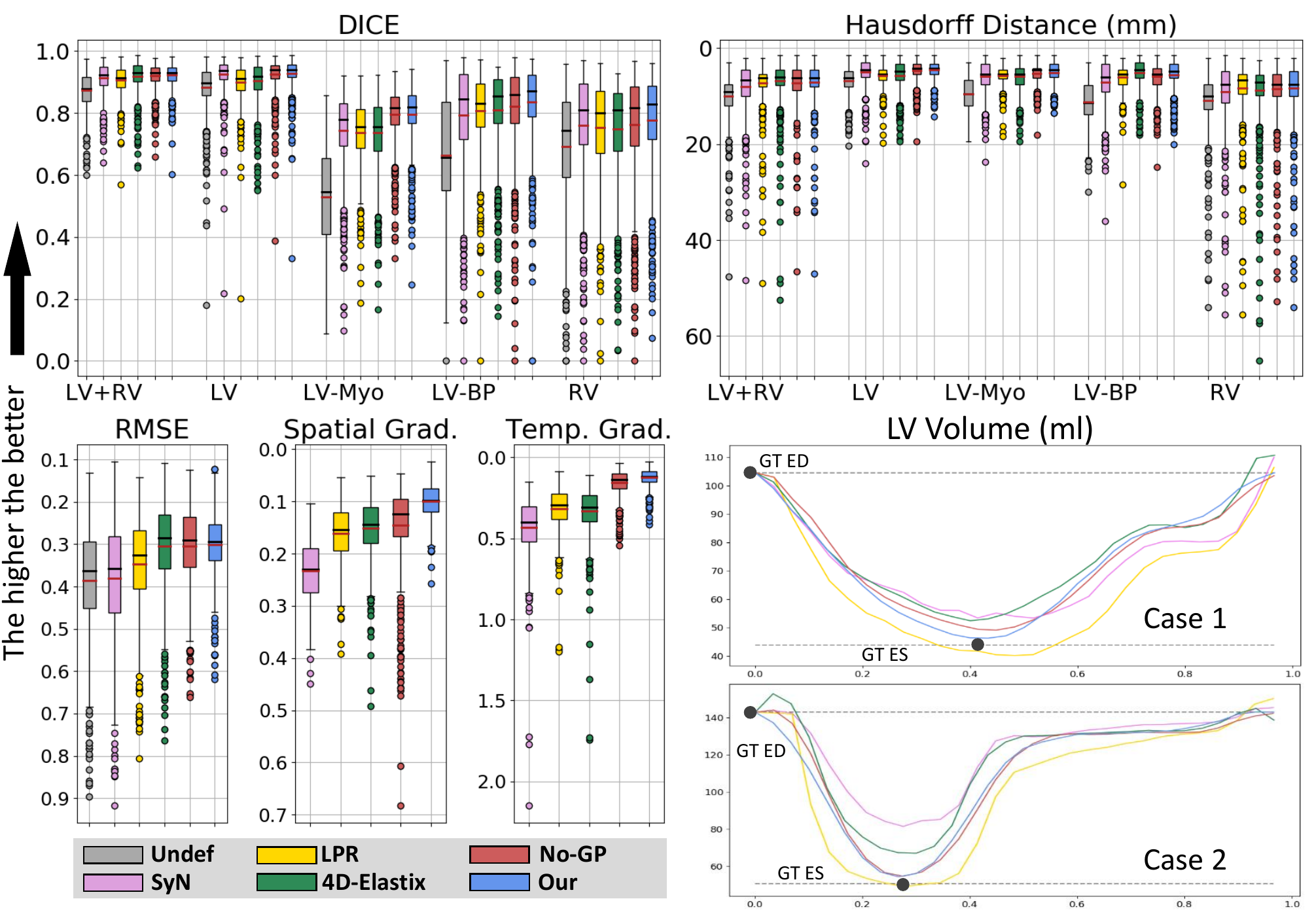}
\caption{\small{Tracking results showing RMSE, spatial and temporal gradients of the displacement fields, DICE scores and  Hausdorff distances for all 2D+t test sequences. The LV volume curves extracted from the warped ED blood pool masks for 2 random test cases in ml, show the temporal smoothness and the distance to the ground-truth ED and ES volumes (marked with black points). The proposed algorithm (Our) shows slightly higher registration accuracy and temporally smoother deformations than the state-of-the-art algorithms: SyN \cite{avants2008symmetric}, LPR \cite{krebs2019learning}, 4D-Elastix \cite{metz2011nonrigid} and the previous version of our method without GP prior (No-GP \cite{krebs2019probabilistic}).}}\label{motion:reg_results}
\end{figure*}

\begin{figure*}[tb]
\centering
\includegraphics[trim=0 135 462 11,clip,width=1.\linewidth]{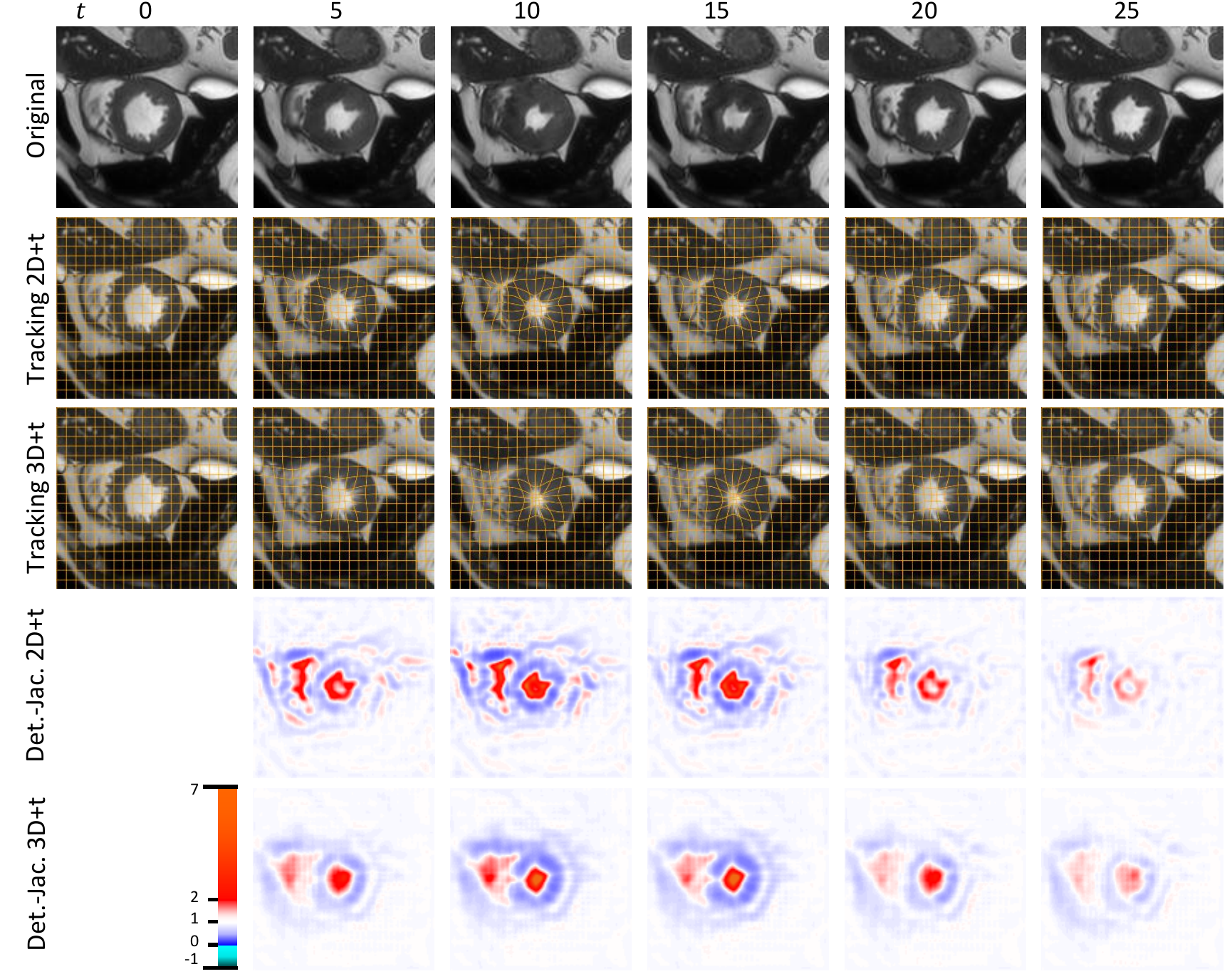}
\caption{\small{Showing 2D+t and 3D+t tracking results of the warped moving image $I_0$ with grid overlay and the Jacobian determinant (Det.-Jac.) \upr{for the mid-ventrical slice of a test sequence}. In 3D+t, smoother Jacobian determinants were obtained.}}\label{motion:reg_images}
\end{figure*}

\subsection{Implementation Details}\label{imp_deetails}
The feature extractor consisted of 4 convolutional layers with (2,2,2,1)-strides and (16,32,32,4)-feature maps and a fully-connected layer of size $2D$, outputting $\gamma_t$. The decoder $p_\theta$ consisted of 3 deconvolutional and 1 convolutional layer with (32,32,32,16)-feature maps. \upd{All (de-)convolutional layers in encoder and decoder used a kernel size of 3. The TCN consisted of four 3x3 1-D convolutional blocks with (1,2,4,8)-dilations. The spatio-temporal Gaussian layer used the spatial $\sigma_G=3$mm and temporal standard deviation $\sigma_T=1.5$, the exponentiation layer used 6 \textit{scaling-squaring} iterations computed using the formula in \cite{arsigny2006log}}. The latent dimensionality was set to $D=32$ (as in \cite{krebs2019learning}). We set the sequence length $T$ to 35, the maximum sequence length found in the training data, resulting in a motion matrix $z$ with $D\cdot \upd{T}=1088$ elements. \upr{The frames of shorter sequences were evenly distributed over $T$ time steps and the gaps were marked as missing data} as described in section \ref{motion:missingdata}. The number of trainable parameters $(\omega,\theta)$ in the network summed up to $\sim$210k in \upd{2D+t and $\sim$456k in 3D+t respectively. L2 weight decay of $1\cdot 10^{-4}$ was applied on all layers} The Cauchy-kernel parameters were chosen as proposed in \cite{fortuin2019multivariate} with $l=7$ and $\sigma_K=1.005$. The variance of the data likelihood was set as the variance of intensity residuals of a few well-registered image sequences with $\sigma_L=0.0045$ in 2D+t and $0.00021$ in 3D+t respectively. \upr{For further details of the neural  network architecture, the reader is referred to Appendix C.}

For training, we used a first-order gradient-based method for stochastic optimization (Adam \cite{kingma2014adam}) with a batch size of one and fixed learning rate of 0.00015. The TD probability $\delta$ was 0.5. Random sub-sequence training was only applied for 3D+t with $\mathcal{T}=18$. Online data augmentation containing randomly shifted, rotated, scaled and mirrored images has been applied \upd{to increase generalizability of the model}. The model was implemented using Keras \cite{chollet2015keras} and Tensorflow \cite{abadi2016tensorflow}. The training time was $\sim$15h in 2D+t and 7 days for 3D+t sequences on a NVIDIA GTX TITAN X GPU. 

\subsection{Registration and Motion Prediction}\label{motion:register}
We compare our model in terms of registration accuracy and spatio-temporal deformation regularity with 3 state-of-the-art diffeomorphic methods: SyN \cite{avants2008symmetric}, the learning-based probabilistic pairwise registration (LPR \cite{krebs2019learning}) and the temporal B-spline algorithm in elastix (4D-Elastix \cite{metz2011nonrigid}). We also compare with the previous version of our method \upr{without} Gaussian process prior (No-GP \cite{krebs2019probabilistic}). SyN and 4D-Elastix have been manually tuned on a few training images following the recommendations in the original papers. The LPR algorithm has been trained on a 2D single scale version using all image pairs of a sequence instead of only the end-diastolic/end-systolic (ED, ES) pairs. We measured registration accuracy using the root mean square error (RMSE) of intensities and segmentation-based DICE scores and 95\%-tile Hausdorff distances (HD, in mm) on the five anatomical structures available in ACDC: left ventricle myocardium (LV-Myo), epicardium (LV), left ventricle bloodpool (LV-BP), right ventricle (RV) and LV+RV. In terms of registration regularity, we report spatial (Spatial
Grad.) and temporal gradients (Temp. Grad.) of the deformation fields $\phi_t$ with $t\in [1,T]$.

\begin{table}[ht]
\centering
\caption{\small{Registration performance with mean and standard deviation scores of DICE (in \%), Hausdorff Distance (HD in mm), spatial and temporal gradients of the deformation fields ($\times 10^{-2}$) comparing our method with the undeformed case (Und), SyN, learning-based pairwise registration (LPR), 4D-Elastix (4D-E) and our previous version without GP prior (No-GP) in 2D+t.}}\label{motion:resultTable}
\begin{tabular}{l|rrrrr}
Method & DICE & HD & Spat. Grad. & Temp. Grad. \\ 
\hline
Und & 72.8 $\pm$14 & 9.70 $\pm$4.20 & -- & --\\ 
SyN   & 82.7 $\pm$12 & 7.02 $\pm$4.34 & 0.23 $\pm$0.06 & 0.43 $\pm$0.19\\ 
LPR   & 82.1 $\pm$10 & 6.60 $\pm$3.07 & 0.16 $\pm$0.06 & 0.32 $\pm$0.13\\ 
4D-E   & 83.7 $\pm$11 & 6.27 $\pm$3.91 & 0.15 $\pm$0.06 & 0.33 $\pm$0.15\\ 
No-GP  & 84.6 $\pm$10 & 6.24 $\pm$3.30 & 0.14 $\pm$0.08 & 0.15 $\pm$0.08\\ 
Our   & 85.2 $\pm$09 & 6.11 $\pm$3.28 & 0.10 $\pm$0.03 & 0.12 $\pm$0.05\\ 
\noalign{\smallskip}
\hline
\end{tabular}
\end{table}

The reported results in Table~\ref{motion:resultTable} were measured on all 2D test sequences containing at least one mask (resulting in 677 sequences from 100 test subjects). DICE scores and Hausdorff distances are only reported for the frames with available ground-truth segmentation (ES images). Detailed box plots of the results together with LV volume curves are shown in Fig.~\ref{motion:reg_results}. The LV volumes (in ml) were extracted by warping the ED mask according to the extracted deformation fields and computing the blood pool volume for all slices of one subject over time. The results indicate that our model achieves the same (RMSE) or slightly better (DICE and HD) registration accuracy compared to the reference methods while improving spatial and temporal regularity as shown by the deformation field gradients and the volume curves. 

\begin{table}[ht]
\centering
\caption{\small{3D+t registration performance with mean and standard deviation scores of RSME, DICE, Hausdorff Distance (HD), spatial and temporal gradients of the deformation fields comparing the undeformed case (Und), 4D-Elastix (4D-E) and the proposed method.}}\label{motion:resultTable3d}
\begin{tabular}{l|rrrrr}
  & DICE & HD & Spat. G. & Temp. G. \\ 
\hline
Und & 70.1 $\pm$12 & 7.7 $\pm$2.7 & -- & --\\ 
4D-E & 79.2 $\pm$10 & 5.1 $\pm$2.1 & 0.15 $\pm$0.06 & 0.62 $\pm$0.32\\ 
Our   & 79.5 $\pm$09 & 5.4 $\pm$2.1 & 0.07 $\pm$0.02 & 0.09 $\pm$0.03\\ 
\noalign{\smallskip}
\hline
\end{tabular}
\end{table}

In Table \ref{motion:resultTable3d}, we show the results on the 100 test sequences for our 3D+t model. In comparison to 4D-Elastix, our 3D+t model shows a similar registration accuracy but a significantly improved spatial and temporal regularity. In addition, our model has a lower RMSE with 0.16 $\pm$ 0.05 compared to 4D-Ealstix with 0.18 $\pm$ 0.07. In Fig.~\ref{motion:reg_images}, the warped moving image $I_0$ and the Jacobian determinant are visualized for one test sequence in 2D+t and 3D+t. One can see, the Jacobian determinants are smoother in 3D+t compared to 2D+t sequences. 

\begin{figure}[tb]
\centering
\includegraphics[trim=28 239 657 7,clip,width=1.\linewidth]{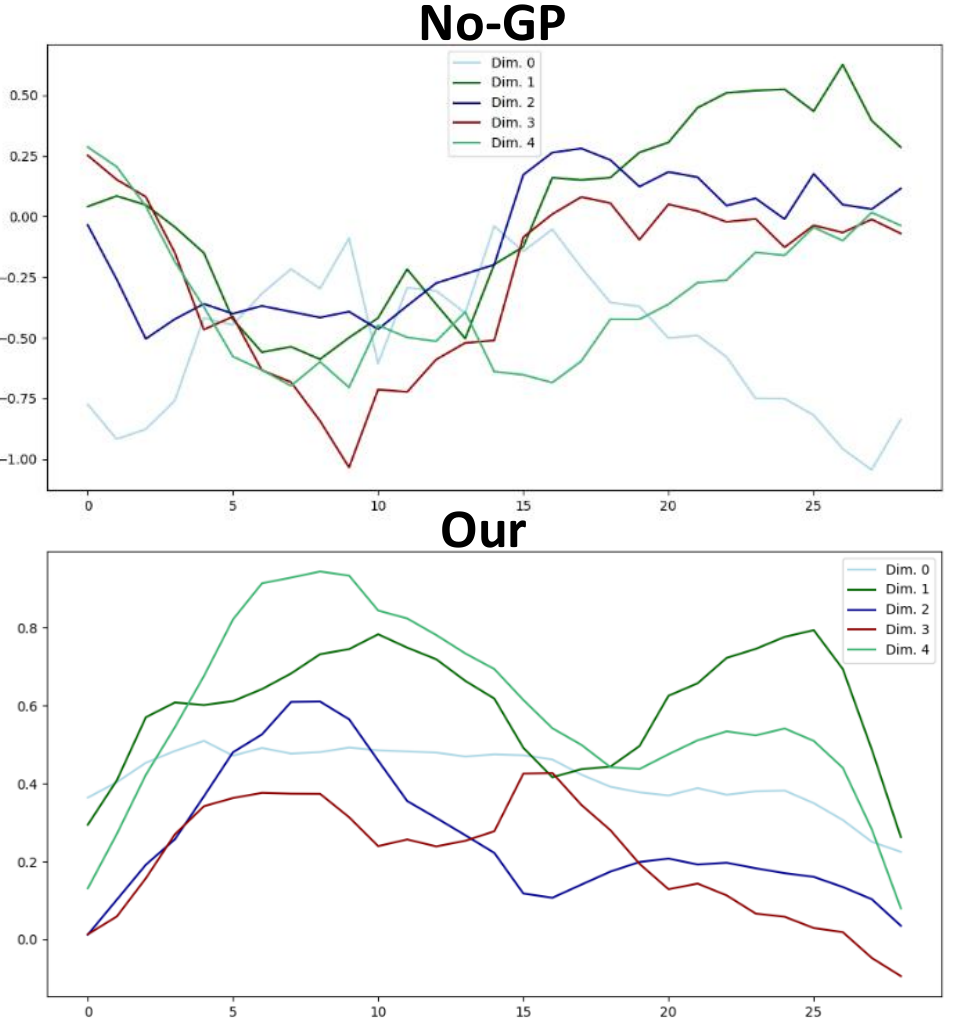}
\caption{\small{First 5 latent dimensions of the same test sequence shows a temporally smoother motion matrix $z$ \upr{(sampled from posterior given $\mu$ and $\sigma$)} for the proposed model trained with the Gaussian process prior compared to the No-GP version.}}\label{motion:latent_space}
\end{figure}

The new Gaussian process prior leads to smoother deformations compared to the previous time-independent prior (No-GP version) while using the same deformation field regularizer. This can be also seen in Fig.~\ref{motion:latent_space} where the first 5 latent dimensions, the sequences \upr{$z_{d\cdot}$} with $d\in[0,4]$, are visualized for one test case. \upd{Furthermore, we investigated the insensitivity of our motion model with respect to initial alignments of the test sequences with the motion model. To this end, we rotated all test sequences by 0, 90, 180 and 270 degrees and compared the performance in Appendix D. We found no statistical significant differences between the results of the 4 test runs demonstrating the orientation independence of the motion model.}

\subsection{Motion Simulation, Interpolation and Transport}

\begin{figure}[tb]
\centering
\includegraphics[trim=15 65 652 0,clip,width=1.\linewidth]{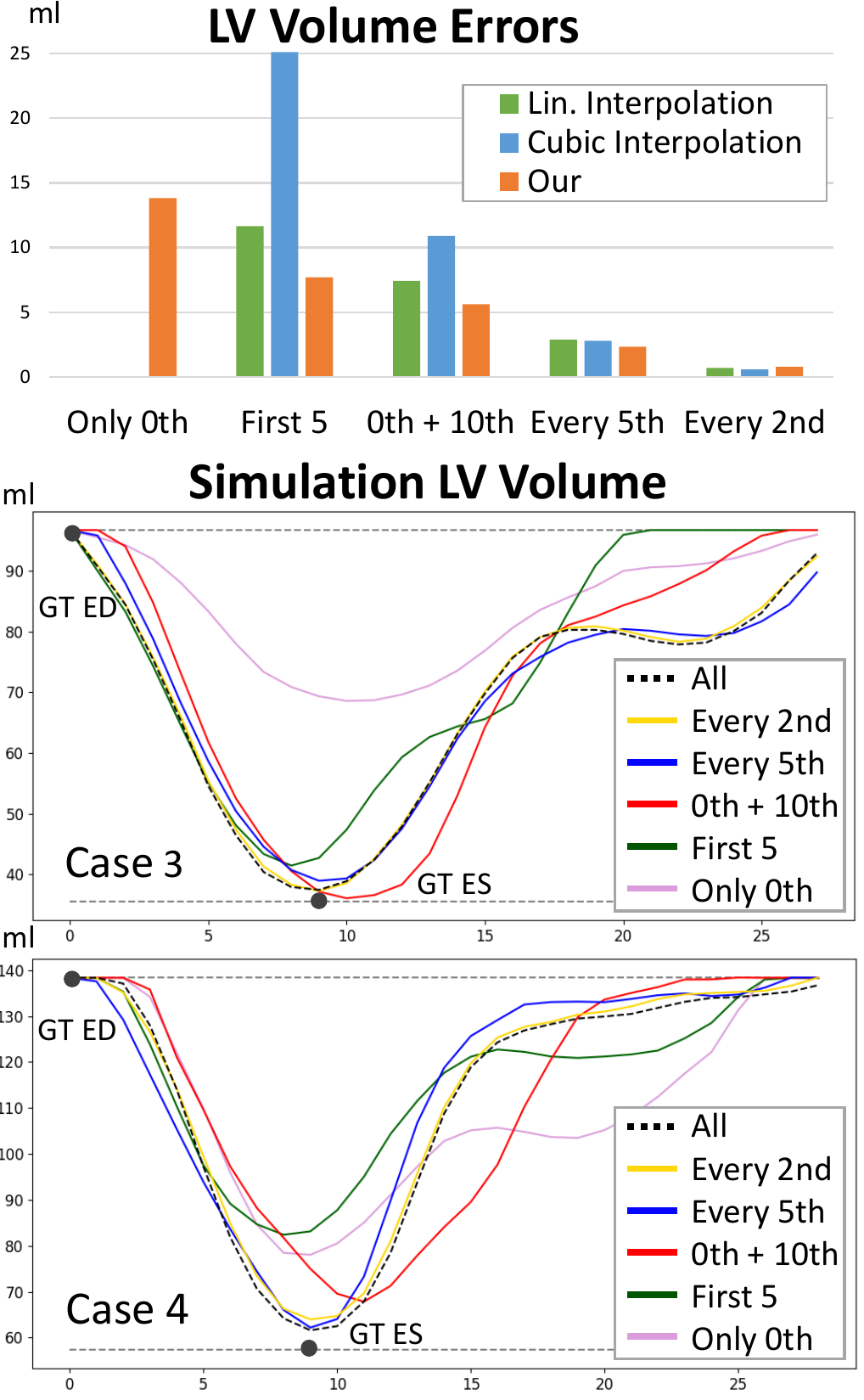}
\caption{\small{Predicted simulated and interpolated motion from a limited number of frames. Provided frames are decreasing from all frames to only the 0th frame (full motion simulation). The volume errors with respect to the all frame prediction are compared with linear and cubic interpolation of the deformation fields. Two random test subjects are shown in the bottom.}}\label{motion:simu_results}
\end{figure}

\begin{figure}[tb]
\centering
\includegraphics[trim=325 170 188 7,clip,width=1\linewidth]{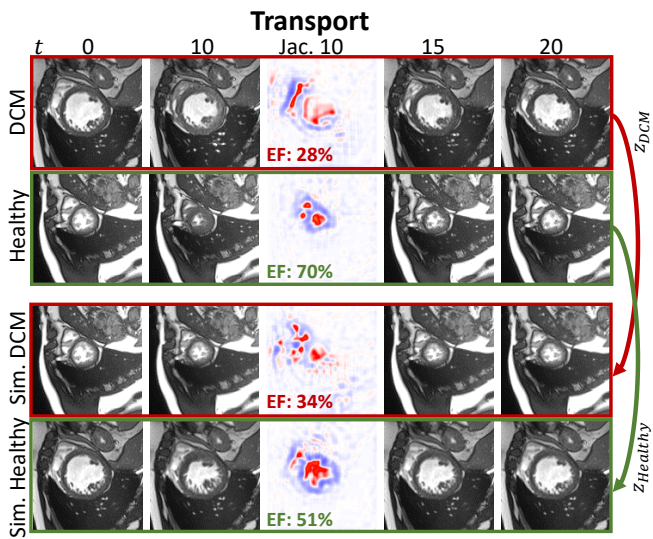}
\caption{\small{Transporting the motion matrix $z$ from one subject and combining it with the end-diastolic frame of another subject allows for simulating a disease (dilated myopathy, DCM, red motion) in a healthy subject and vice versa (green motion). Ejection fraction (EF) of the simulated cases \upr{(Sim. DCM and Sim. Healthy) are more similar to the original transported motion}.}}\label{motion:transport}
\end{figure}

To evaluate the performance on motion interpolation and simulation, we challenged our model to predict the motion for all time steps from a limited number of input frames. Thus, the goal was to predict motion patterns that are as close as possible to the observed motion of the full sequence (i.e.~all registered frames obtained in the all frame model of the previous section \ref{motion:register}). Just as in temporal dropout during training, all the missing frames were represented as zero columns $\gamma_t$ in the feature matrix $\Gamma$ as shown in Fig.~\ref{motion:td}. We compared the motion predictions from various input frame subsets that are provided to the model. First, we provided every 2nd or every 5th frame for motion interpolation. Then, we provided the first 5 frames or only the 10th frame (0th + 10th) to see if the model is able to complete typical cardiac motion patterns. Finally, we tested the full motion simulation by letting the model find a motion sequence given only the moving image $I_0$ (only 0th) and setting feature matrix $\Gamma$ to zero everywhere. We compared the simulated motion, with linear and cubic interpolation of the deformation fields (which are taken from the all frame model at the selected time steps). In the top of Fig.~\ref{motion:simu_results}, average LV volume errors (RMSE) with respect to the all frame model were computed for all 677 test sequences in comparison to linear and cubic interpolation. In the bottom of Fig.~\ref{motion:simu_results}, one can see the results of our model for the different interpolation cases in terms of LV volume curves for two example sequences. 

For the cases of providing every 2nd and every 5th frame, our model interpolated the motion similarly well as linear or cubic interpolation, while providing better results in the cases of providing the 0th+10th and first 5 frames signaling an improved learned cardiac motion model. The full simulation (only 0th) did not result in well fitted volume curves, which is expected as the model has to simulate the full motion sequence from just the ED frame. However, it is observable that the model learned realistic cardiac specific motion patterns as the volume curves for example show the plateau phase before atrial systole which can be also seen in the completed motion for the cases where we provide the first 5 and 0th+10th frames. For the full simulation, our model often slightly under-estimated the motion (cf.~case 3 in Fig.~\ref{motion:simu_results}) which can be related to the pathology distribution in the training dataset which contained many cases with reduced cardiac motion.

Furthermore, we demonstrate the model's capacity of motion transport in a qualitative way. Our model allows to transport motion patterns from one subject to another by taking the motion matrix $z$ of one case and applying it on the moving image of another image sequence (ED frame). In this way, for example a pathological motion can be simulated in a healthy subject or vice versa. In Fig.~\ref{motion:transport}, we present 2 subjects from the ACDC dataset, from which one is classified as healthy and the other as a dilated myopathy case (DCM). We extracted the motion matrices for both and applied them on the ED frame of the other case, such that we simulated a DCM typical motion in the healthy case while \textit{curing} the pathological case. This can be seen for example from the LV contraction strengths in the Jacobian determinants or the related ejection fraction (EF). Note, that this form of parallel transport does not require any additional inter-subject registration.

\section{Discussion}


Our approach has shown state-of-the art registration accuracy and improved deformation regularity temporally and spatially in comparison to 3 state-of-the-art algorithms indicating that the low-dimensional motion encoding helps to regularize the registration problem of \upd{cardiac} image sequences. We have shown that the novel Gaussian process prior leads to a higher temporal consistency compared to the time-independent prior \cite{krebs2019probabilistic} both, in latent and deformation space. A temporally smoother latent space is desirable as it brings more structure and interpretability and is consistent with the temporally smooth motion we experience in deformation space. We have demonstrated motion simulation and interpolation from a very limited number of frames indicating that data acquisition could be speed up as fewer frames are required in order to retrieve an accurate motion. In case of full simulations, our model showed a slightly reduced cardiac motion compared to healthy subjects. The authors believe this is due to a bias introduced from the disease distribution in the training data. To not end up with such a mean motion that merges several pathological motion patterns, one could think of generating disease-specific models. This could be achieved by training different motion models with training sets separated by diseases. 
\upd{We assumed intensities to be constant within an image sequence by using a Gaussian log-likelihood distribution (SSD criterion). Contrast variations can be handled for example by deploying a local cross-correlation distribution as in \cite{krebs2019learning}. However, we found that in the given use case the SSD criterion worked slightly better. Thus, we chose the Gaussian likelihood distribution in this work.}
As another extension to our previous work, we have shown first results on 3D+t sequences which showed smoother Jacobian determinants than the 2D+t version which can be explained by out-of-plane deformations \upr{and the fact that we kept the latent dimensionality $D$ the same for 2D and 3D versions of our algorithm. As the full deformation fields in 3D have more parameters than in 2D the results reconstructed from latent parameters are more smoothed. A lower amount of smoothness and more deformation details could be reached by increasing the latent dimensionality $D$.} However, a limitation is the high computational costs for 3D+t sequences with long training times even for relatively low-dimensional images. 

In future work, we aim to reduce this complexity and \upd{show the approach's potential generalizability to other applications such as for example respiratory motion estimation in dynamic images of the lung. A natural limitation of the proposed model is its low-dimensional latent representation that could become a bottleneck when facing more difficult deformation patterns in other use cases. Also, if the experienced variance in motion in the training data becomes larger, the model first requires more training data  but may also encounter difficulties in finding a reasonable latent representation that is able to capture all the variations in the training data.} 

\upd{\section{Conclusion}}
We presented a probabilistic motion model \upd{learned from images} that can be useful for spatio-temporal registration, temporal super-resolution, data augmentation, shorter acquisition times and motion analysis \upd{for cardiac cine-MRI}. Based on a novel Gaussian process prior conditional variational autoencoder, the model \upd{captures intrinsic motion patterns encoded in a low-dimensional probabilistic space -- the motion matrix. We have shown that such a space allows for accurate diffeomorphic tracking, temporal interpolation, motion simulation and motion transport. The authors believe the presented application-specific motion model that does not rely on hand-crafted features such as bio-mechanical parameters could help in the understanding and analysis of moving organs such as the heart. The results indicate that it is possible to extract a small number of latent parameters in an unsupervised fashion to describe the cardiac motion without requiring much pre-processing of the image sequences. Furthermore, the authors believe the motion matrix as a compact representation of organ motion can be helpful as a quantitative new tool to guide the diagnosis, prognosis or therapy of diseases of dynamic organs such as the heart.} \\


%
\noindent \textbf{Disclaimer: } The concepts and information presented in this paper are based on research results that are not commercially available.

\ifCLASSOPTIONcaptionsoff
  \newpage
\fi

\section*{Appendix}
\subsection{KL Divergence using the GP Prior}
Given 2 multivariate Gaussian distributions with the same dimensionality, the KL divergence is defined in \cite{duchi2007derivations}. Suppose, we take our prior distribution $p(z)$ with zero-mean $\mathbf{0}$ and covariance $\Sigma$ of the form of \ref{motion:cov_prior} and our posterior distribution $q_\omega$ with mean $\mu$ and covariance $\Sigma^*$ with dimensionality $D\upd{T}$:
\begin{multline}\label{motion:kl_gauss}
\text{KL}[q_\omega(z|I_{0:T}) \| p(z)] = \\ \frac{1}{2}\bigg(\text{tr}(\Sigma^{-1}\Sigma^*) + \mu^\top \Sigma^{-1}\mu -D\upd{T} + \ln{\Big(\frac{\det{\Sigma}}{\det{\Sigma^*}} \Big)} \bigg).
\end{multline}
The determinants of the block diagonal matrices $\Sigma$, $\Sigma^*$ are $\det{\Sigma}=|K|^D$ and $\det{\Sigma^*}=|K|^D\prod_{i=1}^D\sigma_i^2$. Thus, the logarithm of the fraction of determinants in \ref{motion:kl_gauss} becomes:
\begin{equation}
\ln{\Big(\frac{\det{\Sigma}}{\det{\Sigma^*}}\Big)} = \ln{\bigg(\frac{1}{\prod_{i=1}^D\sigma_i^2}\bigg)} = -\sum_{i=1}^D \ln{\sigma_i^2}
\end{equation}

When taking the sum over the $D$ latent dimensions over the remaining terms, \ref{motion:kl_gauss} simplifies to:
\begin{equation}\label{motion:kl_loss}
\text{KL}[q_\omega(z|I_{0:T}) \| p(z)] = 
\frac{1}{2}\sum_{i=1}^D \sigma_i^2 \upd{T} + \bar{\mu}_i^\top K^{-1}\bar{\mu}_i - \upd{T} - \ln{(\sigma_i^2)}
\end{equation}
with $\bar{\mu}_i$ being the $i$-th segment of length $T$ in $\mu$. In the case of prior and posterior being identical, thus $\mu=\mathbf{0}$ and $\sigma=\mathbf{1}$ the quantity in \ref{motion:kl_loss} becomes $0$.

\subsection{Cholesky Decomposition of $\Sigma^*$}
The Cholesky decomposition of a symmetric positive-definite matrix $X$ equals the matrix product of a lower-diagonal $L$ and its transposed: $X=LL^\top$. The entries of $L$ can be computed by the Cholesky-Banachiewicz algorithm:
\begin{multline}\label{motion:cholesky}
L_{j,j}=\sqrt{X_{j,j}-\sum_{k=1}^{j-1}L_{j,k}^2} \\
L_{i,j}=\frac{1}{L_{j,j}}\bigg(X_{i,j} - \sum_{k=1}^{j-1} L_{i,k} L{j,k} \bigg) \; \; \; \; \; \; \; \text{for } i>j.
\end{multline}
In case of the block diagonal matrix $\Sigma^*$ the lower triangular matrix $L^*$ equals a block diagonal matrix with lower triangular matrices that are resulting from the Cholesky decompositions of the diagonal block elements of $\Sigma^*$. Thus, in order to compute $L^*$, the Cholesky decompositions of the $i\in D$ diagonal elements $\sigma_i K$ must be computed. From Eq.~\ref{motion:cholesky} it follows that $c\cdot X = (\sqrt{c}\cdot L)(\sqrt{c}\cdot L^\top)$. Thus, $\sigma_i K=(\sqrt{\sigma_i}\cdot L_K)(\sqrt{\sigma_i}\cdot L_K^\top)$ and $L^*$ is:
\begin{equation}
L^* = \text{Diag}_{d=1}^D (\sqrt{\sigma_d}\cdot L_K ) .
\end{equation}
Since the kernel matrix $K$ is fixed in our framework, $L_K$ can be pre-computed using \ref{motion:cholesky} and reused keeping the computational efforts minimal even for a large covariance matrix $\Sigma^*$.

\begin{figure}[tb]
\centering
\includegraphics[trim=3 0 491 0,clip,width=1\linewidth]{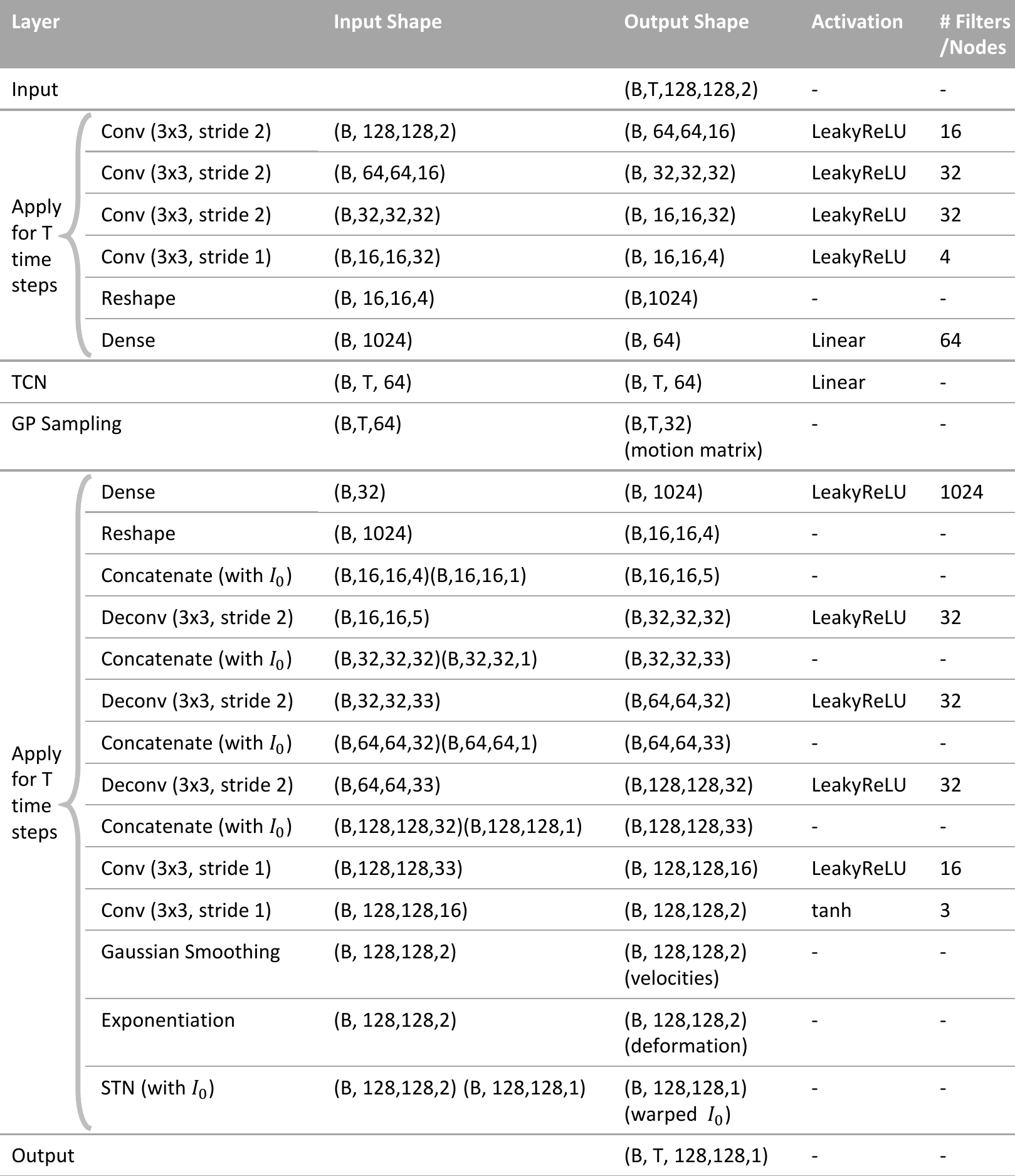}
\upr{\caption{\small{Summary of the neural network architecture of the 2D version of the presented algorithm. Note that most layers are shared over time and are applied on all time instances $T$ with shared weights. The batch size is denoted with B.}}}\label{motion:architecture}
\end{figure}

\upr{\subsection{Network architecture}
The neural network architecture is presented in Fig.~\ref{motion:architecture}. The presented configuration was used for the 2D version of the proposed method. The architecture in 3D is identical just that the 3rd dimension with size 18 is added, thus using 3D convolutional respectively deconvolutional layers. This presents one possible architecture, other, for example deeper network are possible likewise.
}

\subsection{Alignment Sensitivity}
\begin{table}[H]
\centering
\caption{\upd{\small{Registration performance comparing the test data set rotated counter-clockwise with 0, 90, 180 and 270 degrees respectively (in 2D-t). The network has not been retrained. The last row shows average and standard deviations (summary) of the four test runs.}}}\label{motion:resultsAlignment}
\begin{tabular}{l|rrrrr}
 & \upd{DICE} & \upd{HD} & \upd{Spat. G.} & \upd{Temp. G.} \\ 
\hline
\upd{$0^\circ$} & \upd{85.2 $\pm$09} & \upd{6.11 $\pm$3.28} & \upd{0.10 $\pm$0.03} & \upd{0.13 $\pm$0.05}\\
\upd{$90^\circ$} & \upd{84.9 $\pm$09} & \upd{6.24 $\pm$3.27} & \upd{0.10 $\pm$0.03} & \upd{0.13 $\pm$0.05}\\ 
\upd{$180^\circ$} & \upd{85.1 $\pm$09} & \upd{6.08 $\pm$3.25} & \upd{0.11 $\pm$0.04} & \upd{0.13 $\pm$0.06}\\ 
\upd{$270^\circ$} & \upd{85.0 $\pm$09} & \upd{6.10 $\pm$3.19} & \upd{0.11 $\pm$0.04} & \upd{0.13 $\pm$0.06}\\ 
\hline
\upd{summary} & \upd{85.1 $\pm$0.11} & \upd{6.14 $\pm$0.07} & \upd{0.11 $\pm$0.002} & \upd{0.13 $\pm$0.004}\\ 
\noalign{\smallskip}
\end{tabular}
\end{table}

\hspace{1cm}
\bibliography{IEEEabrv.bib,TMI-2020-1689R1.bib}

\begin{thebibliography}{10}
\providecommand{\url}[1]{#1}
\csname url@samestyle\endcsname
\providecommand{\newblock}{\relax}
\providecommand{\bibinfo}[2]{#2}
\providecommand{\BIBentrySTDinterwordspacing}{\spaceskip=0pt\relax}
\providecommand{\BIBentryALTinterwordstretchfactor}{4}
\providecommand{\BIBentryALTinterwordspacing}{\spaceskip=\fontdimen2\font plus
\BIBentryALTinterwordstretchfactor\fontdimen3\font minus
  \fontdimen4\font\relax}
\providecommand{\BIBforeignlanguage}[2]{{%
\expandafter\ifx\csname l@#1\endcsname\relax
\typeout{** WARNING: IEEEtran.bst: No hyphenation pattern has been}%
\typeout{** loaded for the language `#1'. Using the pattern for}%
\typeout{** the default language instead.}%
\else
\language=\csname l@#1\endcsname
\fi
#2}}
\providecommand{\BIBdecl}{\relax}
\BIBdecl

\bibitem{bernard2018deep}
O.~Bernard \emph{et~al.}, ``Deep learning techniques for automatic {MRI}
  cardiac multi-structures segmentation and diagnosis: Is the problem solved?''
  \emph{IEEE Transactions on Medical Imaging}, vol.~37, no.~11, pp. 2514--2525,
  2018.

\bibitem{girija20174d}
J.~Girija, G.~K. Murthy, and P.~C. Reddy, ``4d medical image registration: A
  survey,'' in \emph{2017 International Conference on Intelligent Sustainable
  Systems (ICISS)}.\hskip 1em plus 0.5em minus 0.4em\relax IEEE, 2017, pp.
  539--547.

\bibitem{rohe2018low}
M.-M. Roh{\'e}, M.~Sermesant, and X.~Pennec, ``Low-dimensional representation
  of cardiac motion using barycentric subspaces: A new group-wise paradigm for
  estimation, analysis, and reconstruction,'' \emph{Medical Image Analysis},
  vol.~45, no.~1, pp. 1--12, 2018.

\bibitem{sotiras}
A.~Sotiras, C.~Davatzikos, and N.~Paragios, ``Deformable medical image
  registration: A survey,'' \emph{IEEE Transactions on Medical Imaging},
  vol.~32, no.~7, pp. 1153--1190, 2013.

\bibitem{vercauteren2009diffeomorphic}
T.~Vercauteren, X.~Pennec, A.~Perchant, and N.~Ayache, ``Diffeomorphic demons:
  Efficient non-parametric image registration,'' \emph{NeuroImage}, vol.~45,
  no.~1, pp. S61--S72, 2009.

\bibitem{peyrat2010registration}
J.-M. Peyrat, H.~Delingette, M.~Sermesant, C.~Xu, and N.~Ayache, ``Registration
  of 4d cardiac ct sequences under trajectory constraints with multichannel
  diffeomorphic demons,'' \emph{IEEE Transactions on Medical Imaging}, vol.~29,
  no.~7, pp. 1351--1368, 2010.

\bibitem{beg2005computing}
M.~F. Beg, M.~I. Miller, A.~Trouv{\'e}, and L.~Younes, ``Computing large
  deformation metric mappings via geodesic flows of diffeomorphisms,''
  \emph{International Journal of Computer Vision}, vol.~61, no.~2, pp.
  139--157, 2005.

\bibitem{zhang2015finite}
M.~Zhang and P.~T. Fletcher, ``Finite-dimensional lie algebras for fast
  diffeomorphic image registration,'' in \emph{International Conference on
  Information Processing in Medical Imaging}.\hskip 1em plus 0.5em minus
  0.4em\relax Springer, 2015, pp. 249--260.

\bibitem{vercauteren2008symmetric}
T.~Vercauteren, X.~Pennec, A.~Perchant, and N.~Ayache, ``Symmetric log-domain
  diffeomorphic registration: A demons-based approach,'' in \emph{International
  Conference on Medical Image Computing and Computer-Assisted
  Intervention}.\hskip 1em plus 0.5em minus 0.4em\relax Springer, 2008, pp.
  754--761.

\bibitem{avants2008symmetric}
B.~B. Avants, C.~L. Epstein, M.~Grossman, and J.~C. Gee, ``Symmetric
  diffeomorphic image registration with cross-correlation: evaluating automated
  labeling of elderly and neurodegenerative brain,'' \emph{Medical Image
  Analysis}, vol.~12, no.~1, pp. 26--41, 2008.

\bibitem{lorenzi2013lcc}
M.~Lorenzi, N.~Ayache, G.~B. Frisoni, and for~the Alzheimer's Disease
  Neuroimaging Initiative~(ADNI), ``{LCC-D}emons: a robust and accurate
  symmetric diffeomorphic registration algorithm,'' \emph{NeuroImage}, vol.~81,
  no.~1, pp. 470--483, 2013.

\bibitem{yang2017quicksilver}
X.~Yang, R.~Kwitt, M.~Styner, and M.~Niethammer, ``Quicksilver: Fast predictive
  image registration--a deep learning approach,'' \emph{NeuroImage}, vol. 158,
  no.~1, pp. 378--396, 2017.

\bibitem{rohe2017svf}
M.-M. Roh{\'e}, M.~Datar, T.~Heimann, M.~Sermesant, and X.~Pennec, ``{SVF}-net:
  Learning deformable image registration using shape matching,'' in
  \emph{International Conference on Medical Image Computing and
  Computer-Assisted Intervention}.\hskip 1em plus 0.5em minus 0.4em\relax
  Springer, 2017, pp. 266--274.

\bibitem{dalca2018unsupervised}
A.~V. Dalca, G.~Balakrishnan, J.~Guttag, and M.~R. Sabuncu, ``Unsupervised
  learning for fast probabilistic diffeomorphic registration,'' in
  \emph{International Conference on Medical Image Computing and
  Computer-Assisted Intervention}.\hskip 1em plus 0.5em minus 0.4em\relax
  Springer, 2018, pp. 729--738.

\bibitem{krebs2019learning}
J.~Krebs, H.~Delingette, B.~Mailh{\'e}, N.~Ayache, and T.~Mansi, ``Learning a
  probabilistic model for diffeomorphic registration,'' \emph{IEEE Transactions
  on Medical Imaging}, vol.~38, no.~9, pp. 2165--2176, Sep 2019.

\bibitem{de2019deep}
B.~D. de~Vos, F.~F. Berendsen, M.~A. Viergever, H.~Sokooti, M.~Staring, and
  I.~I{\v{s}}gum, ``A deep learning framework for unsupervised affine and
  deformable image registration,'' \emph{Medical Image Analysis}, vol.~52,
  no.~1, pp. 128--143, 2019.

\bibitem{dalca2019unsupervised}
A.~V. Dalca, G.~Balakrishnan, J.~Guttag, and M.~R. Sabuncu, ``Unsupervised
  learning of probabilistic diffeomorphic registration for images and
  surfaces,'' \emph{Medical Image Analysis}, vol.~57, no.~1, pp. 226--236,
  2019.

\bibitem{arsigny2006log}
V.~Arsigny, O.~Commowick, X.~Pennec, and N.~Ayache, ``A log-euclidean framework
  for statistics on diffeomorphisms,'' in \emph{International Conference on
  Medical Image Computing and Computer-Assisted Intervention}.\hskip 1em plus
  0.5em minus 0.4em\relax Springer, 2006, pp. 924--931.

\bibitem{ledesma2005spatio}
M.~J. Ledesma-Carbayo \emph{et~al.}, ``Spatio-temporal nonrigid registration
  for ultrasound cardiac motion estimation,'' \emph{IEEE Transactions on
  Medical Imaging}, vol.~24, no.~9, pp. 1113--1126, 2005.

\bibitem{vandemeulebroucke2011spatiotemporal}
J.~Vandemeulebroucke, S.~Rit, J.~Kybic, P.~Clarysse, and D.~Sarrut,
  ``Spatiotemporal motion estimation for respiratory-correlated imaging of the
  lungs,'' \emph{Medical Physics}, vol.~38, no.~1, pp. 166--178, 2011.

\bibitem{de2012temporal}
M.~De~Craene \emph{et~al.}, ``Temporal diffeomorphic free-form deformation:
  Application to motion and strain estimation from 3d echocardiography,''
  \emph{Medical Image Analysis}, vol.~16, no.~2, pp. 427--450, 2012.

\bibitem{metz2011nonrigid}
C.~Metz, S.~Klein, M.~Schaap, T.~van Walsum, and W.~J. Niessen, ``Nonrigid
  registration of dynamic medical imaging data using nd+ t b-splines and a
  groupwise optimization approach,'' \emph{Medical Image Analysis}, vol.~15,
  no.~2, pp. 238--249, 2011.

\bibitem{qin2018joint}
C.~Qin \emph{et~al.}, ``Joint learning of motion estimation and segmentation
  for cardiac mr image sequences,'' in \emph{International Conference on
  Medical Image Computing and Computer-Assisted Intervention}.\hskip 1em plus
  0.5em minus 0.4em\relax Springer, 2018, pp. 472--480.

\bibitem{shi2013temporal}
W.~Shi \emph{et~al.}, ``Temporal sparse free-form deformations,'' \emph{Medical
  Image Analysis}, vol.~17, no.~7, pp. 779--789, 2013.

\bibitem{giger2018respiratory}
A.~Giger \emph{et~al.}, ``Respiratory motion modelling using cgans,'' in
  \emph{International Conference on Medical Image Computing and
  Computer-Assisted Intervention}.\hskip 1em plus 0.5em minus 0.4em\relax
  Springer, 2018, pp. 81--88.

\bibitem{giger2019inter}
A.~Giger \emph{et~al.}, ``Inter-fractional respiratory motion modelling from
  abdominal ultrasound: A feasibility study,'' in \emph{International Workshop
  on PRedictive Intelligence In MEdicine}.\hskip 1em plus 0.5em minus
  0.4em\relax Springer, 2019, pp. 11--22.

\bibitem{caballero2017real}
J.~Caballero \emph{et~al.}, ``Real-time video super-resolution with
  spatio-temporal networks and motion compensation,'' in \emph{Proceedings of
  the IEEE Conference on Computer Vision and Pattern Recognition}, 2017, pp.
  4778--4787.

\bibitem{kappeler2016video}
A.~Kappeler, S.~Yoo, Q.~Dai, and A.~K. Katsaggelos, ``Video super-resolution
  with convolutional neural networks,'' \emph{IEEE Transactions on
  Computational Imaging}, vol.~2, no.~2, pp. 109--122, 2016.

\bibitem{sermesant2008toward}
M.~Sermesant \emph{et~al.}, ``Toward patient-specific myocardial models of the
  heart,'' \emph{Heart failure clinics}, vol.~4, no.~3, pp. 289--301, 2008.

\bibitem{yang2011prediction}
L.~Yang, B.~Georgescu, Y.~Zheng, Y.~Wang, P.~Meer, and D.~Comaniciu,
  ``Prediction based collaborative trackers (pct): A robust and accurate
  approach toward 3d medical object tracking,'' \emph{IEEE Transactions on
  Medical Imaging}, vol.~30, no.~11, pp. 1921--1932, 2011.

\bibitem{qiu2011principal}
A.~Qiu, L.~Younes, and M.~I. Miller, ``Principal component based diffeomorphic
  surface mapping,'' \emph{IEEE Transactions on Medical Imaging}, vol.~31,
  no.~2, pp. 302--311, 2011.

\bibitem{jud2015respiratory}
C.~Jud, F.~Preiswerk, and P.~C. Cattin, ``Respiratory motion compensation with
  topology independent surrogates,'' in \emph{Workshop on imaging and computer
  assistance in radiation therapy}, 2015.

\bibitem{jud2017localized}
C.~Jud, A.~Giger, R.~Sandk{\"u}hler, and P.~C. Cattin, ``A localized
  statistical motion model as a reproducing kernel for non-rigid image
  registration,'' in \emph{International Conference on Medical Image Computing
  and Computer-Assisted Intervention}.\hskip 1em plus 0.5em minus 0.4em\relax
  Springer, 2017, pp. 261--269.

\bibitem{kingma2014semi}
D.~P. Kingma, S.~Mohamed, D.~J. Rezende, and M.~Welling, ``Semi-supervised
  learning with deep generative models,'' in \emph{Advances in Neural
  Information Processing Systems}, 2014, pp. 3581--3589.

\bibitem{krebs2019probabilistic}
J.~Krebs, T.~Mansi, N.~Ayache, and H.~Delingette, ``Probabilistic motion
  modeling from medical image sequences: Application to cardiac cine-mri,'' in
  \emph{Statistical Atlases and Computational Models of the Heart.
  Multi-Sequence CMR Segmentation, CRT-EPiggy and LV Full Quantification
  Challenges}.\hskip 1em plus 0.5em minus 0.4em\relax Cham: Springer, 2020, pp.
  176--185.

\bibitem{kingma2013auto}
D.~P. Kingma and M.~Welling, ``Auto-encoding variational bayes,'' \emph{arXiv
  preprint arXiv:1312.6114}, 2013.

\bibitem{kingma2014adam}
D.~P. Kingma and J.~Ba, ``Adam: A method for stochastic optimization,''
  \emph{arXiv preprint arXiv:1412.6980}, 2014.

\bibitem{koltun2018}
S.~Bai, J.~Z. Kolter, and V.~Koltun, ``An empirical evaluation of generic
  convolutional and recurrent networks for sequence modeling,'' \emph{arXiv
  preprint arXiv:1803.01271}, 2018.

\bibitem{rasmussen2003gaussian}
C.~E. Rasmussen, ``Gaussian processes in machine learning,'' in \emph{Summer
  School on Machine Learning}.\hskip 1em plus 0.5em minus 0.4em\relax Springer,
  2003, pp. 63--71.

\bibitem{fortuin2019multivariate}
V.~Fortuin, G.~R{\"a}tsch, and S.~Mandt, ``Multivariate time series imputation
  with variational autoencoders,'' \emph{arXiv preprint arXiv:1907.04155},
  2019.

\bibitem{balakrishnan2018unsupervised}
G.~Balakrishnan, A.~Zhao, M.~R. Sabuncu, J.~Guttag, and A.~V. Dalca, ``An
  unsupervised learning model for deformable medical image registration,'' in
  \emph{Proceedings of the IEEE Conference on Computer Vision and Pattern
  Recognition}, 2018, pp. 9252--9260.

\bibitem{kingma2019introduction}
D.~P. Kingma and M.~Welling, ``An introduction to variational autoencoders,''
  \emph{Foundations and Trends{\textregistered} in Machine Learning}, vol.~12,
  no.~4, pp. 307--392, 2019.

\bibitem{maas2013rectifier}
A.~L. Maas, A.~Y. Hannun, and A.~Y. Ng, ``Rectifier nonlinearities improve
  neural network acoustic models,'' in \emph{Proc. icml}, vol.~30, no.~1, 2013,
  p.~3.

\bibitem{jaderberg2015spatial}
M.~Jaderberg, K.~Simonyan, A.~Zisserman, and K.~Kavukcuoglu, ``Spatial
  transformer networks,'' in \emph{Advances in Neural Information Processing
  Systems}, 2015, pp. 2017--2025.

\bibitem{chollet2015keras}
F.~Chollet, ``Keras,'' \url{https://keras.io}, 2015.

\bibitem{abadi2016tensorflow}
M.~Abadi \emph{et~al.}, ``Tensorflow: Large-scale machine learning on
  heterogeneous distributed systems,'' \emph{arXiv preprint arXiv:1603.04467},
  2016.

\bibitem{duchi2007derivations}
J.~Duchi, ``Derivations for linear algebra and optimization,'' \emph{Berkeley,
  California}, vol.~3, no.~1, pp. 2325--5870, 2007.

\end{thebibliography}

\end{document}